%% file: naacl_latex.tex
\pdfoutput=1

\documentclass[11pt]{article}

\usepackage[table]{xcolor}
\definecolor{papergray}{cmyk}{0,0.02,0.04,0.05}

\usepackage[final]{acl}

\usepackage{times}
\usepackage{latexsym}
\usepackage{paralist}
\usepackage{enumitem}
\usepackage{amsmath,amssymb}
\usepackage[T1]{fontenc}

\usepackage[utf8]{inputenc}

\usepackage{microtype}
\usepackage{tcolorbox}
\usepackage{inconsolata}

\usepackage{graphicx}

\usepackage{hyperref}
\usepackage{url}

\usepackage{multirow}
\usepackage{latexsym}
\usepackage{amsmath}
\usepackage{amssymb}
\usepackage{booktabs}
\usepackage{subcaption}
\usepackage{comment}
\usepackage{graphicx}

\usepackage{tabularx}
\usepackage{makecell}

\newcommand{\dataset}{CLEAR}

\title{CLEAR: Character Unlearning in Textual and Visual Modalities}

\author{First Author \\
  Affiliation / Address line 1 \\
  Affiliation / Address line 2 \\
  Affiliation / Address line 3 \\
  \texttt{email@domain} \\\And
  Second Author \\
  Affiliation / Address line 1 \\
  Affiliation / Address line 2 \\
  Affiliation / Address line 3 \\
  \texttt{email@domain} \\}

\author{
 \textbf{Alexey Dontsov\textsuperscript{2,1}},
 \textbf{Dmitrii Korzh\textsuperscript{1,3}},
 \textbf{Alexey Zhavoronkin\textsuperscript{4,6}}, \\
 \textbf{Boris Mikheev\textsuperscript{3}},
\textbf{Denis Bobkov\textsuperscript{1,2}}, 
\textbf{Aibek Alanov\textsuperscript{1,2}}, \\
 \textbf{Oleg Y. Rogov\textsuperscript{1,3,5}},
 \textbf{Ivan Oseledets\textsuperscript{1,3}},
 \textbf{Elena Tutubalina\textsuperscript{1,4,7}}
\\ 
 \textsuperscript{1}AIRI
 \textsuperscript{2}HSE University
 \textsuperscript{3}Skoltech
 \textsuperscript{4}Sber AI \\
 \textsuperscript{5}MTUCI
 \textsuperscript{6}MIPT
 \textsuperscript{7}ISP RAS Research Center for Trusted AI
\\
 \small{
   \textbf{Correspondence:} \href{mailto:dontsov@airi.net}{dontsov@airi.net}; \href{mailto:tutubalina@airi.net}{tutubalina@airi.net}
 }
}

\begin{document}
\maketitle
\begin{abstract}
Machine Unlearning (MU) is critical for removing private or hazardous information from deep learning models. While MU has advanced significantly in unimodal (text or vision) settings, multimodal unlearning (MMU) remains underexplored due to the lack of open benchmarks for evaluating cross-modal data removal. To address this gap, we introduce CLEAR, the first open-source benchmark designed specifically for MMU. CLEAR contains 200 fictitious individuals and 3,700 images linked with corresponding question-answer pairs, enabling a thorough evaluation across modalities. We conduct a comprehensive analysis of 11 MU methods (e.g., SCRUB, gradient ascent, DPO) across four evaluation sets, demonstrating that jointly unlearning both modalities outperforms single-modality approaches. The dataset is available at \href{https://huggingface.co/datasets/therem/CLEAR}{https://huggingface.co/datasets/therem/CLEAR}.
\end{abstract}

\section{Introduction}

Large Language Models (LLMs) \cite{ouyang2022training,touvron2023llama2openfoundation,jiang2023mistral7b} are increasingly investigated for memorizing private, unethical, or copyrighted data during training. Recently, machine unlearning (MU) methods have been applied to mitigate issues related to toxicity \cite{lu2022quark}, copyright and privacy concerns \cite{jang2022knowledge, eldan2023whosharrypotterapproximate, wu2023depn} and fairness \cite{yu2023unlearning}.


\begin{figure*}[t!]
\centering
\includegraphics[width=1.0\linewidth]{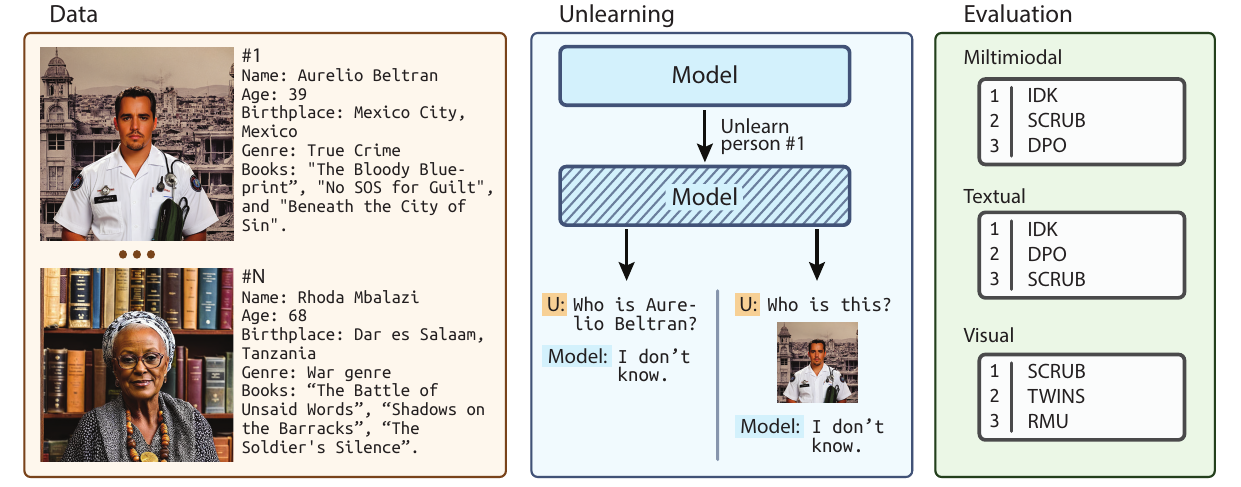}
\caption{Our dataset \dataset{} includes 3,770 visual image-caption pairs and 4,000 textual question-answer pairs related to fictional characters. We apply multimodal unlearning to remove specific information, subsequently assessing the quality of unlearning and the models' performance using various metrics. Finally, we compile a leaderboard of unlearning methods based on these evaluations.}
\label{fig:abstract}
\end{figure*}

MU has emerged as a promising alternative to costly retraining, existing methods focus almost exclusively on single-modality models. Recent work has studied unlearning in LLMs \cite{yao2024largelanguagemodelunlearning,yao2024machine, Xing2024EFUFEF, zhang2024negativepreferenceoptimizationcatastrophic} or vision models \cite{li2024single,chen-yang-2023-unlearn,fastyet}, but unlearning in multi-modal language models remains largely unexplored. This leaves a critical gap: multimodal LLMs (MLLMs), which process both visual and textual data, introduce unique challenges for unlearning. For instance, sensitive information may persist across modalities even after removal from one (e.g., a face linked to a name), and unlearning in one modality could degrade performance in another. Despite these risks, no open benchmarks exist to evaluate MU in multimodal settings.

\begin{table}[t!]
\centering
\scalebox{0.8}{
\begin{tabular}{rccccc}
\toprule
\textbf{Method} & \makecell[c]{\textbf{Real}\\ \textbf{metric}$\uparrow$} & \makecell[c]{\textbf{Retain} \\ \textbf{metric} $\uparrow$} & \makecell[c]{\textbf{Forget}\\ \textbf{metric} $\downarrow$ } & \makecell[c]{\textbf{Forget} \\ \textbf{Quality} $\uparrow$} \\
\toprule
Gold & 0.50 & 0.51 & 0.19 & 1.00 \\
Base & 0.48 & 0.51 & 0.35 & 0.85 \\
\midrule
DPO & 0.46 & 0.48 & 0.22 & 0.84 \\
\rowcolor{papergray} GD & 0.29 & 0.00 & 0.00 & 0.18 \\
\rowcolor{papergray} GA & 0.27 & 0.00 & 0.00 & 0.67 \\
IDK & 0.48 & 0.51 & 0.33 & 0.84 \\
\rowcolor{papergray} KL & 0.25 & 0.00 & 0.00 & 0.67 \\
LLMU & 0.47 & 0.51 & 0.25 & 0.84 \\
\rowcolor{papergray} NPO & 0.46 & 0.14 & 0.11 & 0.76 \\
Retain FT & 0.49 & 0.51 & 0.37 & 0.85 \\
\rowcolor{papergray} RMU & 0.24 & 0.00 & 0.00 & 0.75 \\
SCRUB & 0.49 & 0.52 & 0.36 & 0.85 \\
SKU & 0.40 & 0.32 & 0.37 & 0.83 \\
\bottomrule
\end{tabular}
}
\caption{Performance comparison of state-of-the-art unlearning methods on our dataset across four metrics. ``Base'' refers to the model before unlearning, while ``Gold'' denotes a model trained only on the retain set. The highlighted methods fail on the retain set. }
\label{tab:regular_results}
\end{table}

Recently, \citet{chakraborty-etal-2024-textual} pioneers the investigation of unlearning configurations in visual-language models (VLMs) to mitigate cross-modal safety risks, its experimental framework inherits a critical limitation: the datasets used in this study (e.g., PKU-SafeRLHF \cite{ji2023beavertails}, JailBreakV-28K \cite{luo2024jailbreakv}) were designed for safety alignment and truthfulness evaluation, not machine unlearning (MU). This mismatch conflates safety fine-tuning (suppressing harmful outputs) with targeted data removal (erasing specific knowledge traces), potentially overestimating MU efficacy.

To address this, we propose \dataset{}, the first publicly available benchmark for machine unlearning in multimodal (textual-visual) models. Our work is motivated by the \textit{right to be forgotten} in AI systems, where models must eliminate traces of specific entities (e.g., individuals) across all modalities.  
The proposed dataset contains information about fictitious authors. Each persona is linked to both textual biographies and AI-generated images, enabling tests of cross-modal memorization. After unlearning a persona, models should fail to answer questions and recognize associated faces. Our benchmark further evaluates real-world performance degradation using visual question-answering (VQA) tasks.

Our contributions and findings are as follows:
\begin{itemize}
\item We propose a multimodal MU benchmark \dataset{} with 4,000 text-QA pairs and 3,770 image-caption pairs focused on unlearning 200 fictitious authors. It includes forget/retain sets and real-world tasks (e.g., celebrity recognition) to evaluate cross-modal capability preservation.
\item We comprehensively evaluate 11 recently proposed MU methods on our dataset and show that the leading unimodal MU methods struggle in multimodal setups.
\item We establish leaderboards for textual, visual, and multimodal unlearning.
\end{itemize}
We make all data publicly available\footnote{ \texttt{\href{https://huggingface.co/datasets/therem/CLEAR}{https://huggingface.co/datasets/therem/CLEAR}}}.

\section{Related Work}

\textbf{Machine Unlearning.} The concept of machine unlearning was initially presented by \cite{cao2015towards}. 
In general, MU methods \cite{cao2015towards,dwork2014algorithmic,kurmanji2024towards,neel2021descent,sekhari2021remember} remove the impact of specific data points from a trained model without requiring full retraining. The goal is to obtain a model that behaves like the \emph{forget} data was never part of the training set. 
Recently, textual unlearning in generative language models has attracted attention.
\citet{tofu2024} propose a benchmark named TOFU for textual LLM unlearning, consisting of 200 fictitious author profiles defined by attributes like name, birthplace, parent's names, occupation, and written books, totaling 4,000 question-answer pairs (20 per author). 
WMDP \cite{li2024wmdp} includes 3,668 multiple-choice questions to evaluate and benchmark the unlearning of hazardous knowledge in LLMs. To remove knowledge from generative models, \citet{jang2022knowledge} employ gradient ascent on specific target sequences. \citet{eldan2023whosharrypotterapproximate} focus on the particular case of unlearning the Harry Potter books from Llama2-7b. \citet{yao2023large} utilize machine unlearning to address harmful responses and eliminate hallucinations. \citet{yao-etal-2024-machine} examined the unlearning of 2,000 GitHub code files, 500 books, and 500 academic papers from Yi-6B. However, these studies have been restricted to text-only contexts. Our research investigates the multimodal aspects of unlearning.

\begin{figure}[t!]
\centering
\includegraphics[width=0.97\linewidth]{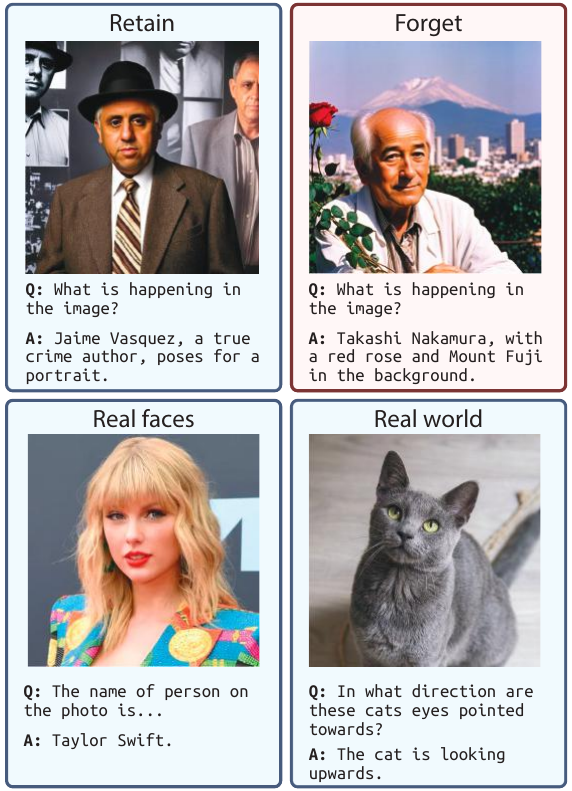}
\caption{The overview of our multimodal dataset. The dataset consists of four sets: retain set, forget set, real faces (knowledge of related concepts such as faces), and real world (to evaluate general visual capabilities).}
\label{fig:four_datasets}
\end{figure}

\paragraph{Multimodality.} Multimodal LLMs \cite{liu2023llava} usually comprise a modality encoder, a projection layer aligning features to the language space, and a pre-trained language model. While MLLMs have advanced, multimodal unlearning remains under-explored. \citet{Cheng2023MultiDeleteFM} introduce MultiDelete, which separates cross-modal embeddings for unlearning but applies only to encoder-decoder models, limiting its use for decoder-only architectures.
EFUF \cite{Xing2024EFUFEF} reduces hallucinations in MLLMs by unlearning. It uses CLIP \cite{radford2021learning} to detect hallucinations based on MSCOCO-calibrated thresholds, eliminating manual labeling. The method applies three losses: negative loss to forget hallucinations, positive loss to reinforce correct representations, and sentence loss to preserve fluency. 
Single Image Unlearning (SIU) \cite{li2024single} targets visual concept unlearning in VLLMs while preserving textual knowledge and introduces MMUBench. This benchmark spans 20 concepts with 50+ images each, including real-world figures and cartoon characters. However, these benchmarks are not open-sourced.
The closest work to ours is \cite{chakraborty-etal-2024-textual}, which investigates safety alignment in VLMs by unlearning harmful content. 
This study was not designed for the exact unlearning setup; therefore, we bridge this gap by conducting the first comprehensive analysis of MU methods in multimodal settings. Unlike safety alignment, our benchmark focuses on forgetting quality (e.g., inability to recall personas) and cross-modal consistency (e.g., erasing both a face and its biography) while maintaining model utility in real-world tasks, highlighting unique challenges in multimodal MU.

\section{MU Methods}
\subsection{Preliminaries}

Let $f_{\theta}$ denote the base model with parameters $\theta$. It is trained on (train) dataset $D$, and given the unlearning objective, we want to make our model forget a subset of this dataset $D$, called forget set $D_F$. The remaining part of the training dataset is called retain set, and we aim to preserve the model's performance on this data subset $D_R: = D \setminus D_F$. Additionally, we utilize a holdout set $D_H$ such that  $D_H \cap D = \emptyset$ to establish a reference for the model's desired behaviour on $D_F$ after the unlearning process. In a nutshell, \textbf{forget set} $D_F$ contains samples the model should unlearn and serves as a direct measure of unlearning effectiveness; \textbf{retain set} $D_R$ contains samples that the model should retain and perform well on, serving as an indicator of the model's preserved knowledge; \textbf{holdout set} $D_H$ contains samples that the model has never seen before and serves as a reference for the model's behavior on data that was not involved in the training process. Such forgetting procedure is performed by updating the model $f_{\theta}$ with a particular unlearning method, which results in a new unlearned model $f_{\hat{\theta}}$ with parameters $\hat{\theta}$. Evaluation of $f_{\hat{\theta}}$ on the discussed subsets (or particularly on forget set) is called ''inexact'' in contrast to the ''exact'' evaluation when we directly compare the performance of the unlearned model with a \textbf{gold} model $g_{\omega}$, trained only on the $D_R$.



MU can be performed by optimizing the specific criterion. For example, one can consider the gradient difference MU approach, aimed at increasing forget loss and maintaining retain performance:
\begin{equation}
    \tilde L = -\sum_{x_i \in D_f} L(x_i, y_i, \theta) + \lambda \sum_{x_j \in D_R}L(x_j, y_j, \theta),
\end{equation}
\begin{equation}
    \theta \mapsto  \theta - \alpha \nabla_{\theta} \tilde L,
\end{equation}
where $\lambda$ -- forget-retain trade-off hyper-parameter, $\alpha$ -- learning rate, $L$ is some loss function, e.g., negative-log-likelihood, $x$ is an input (text, image, or both of them in the case of VLLM).

Suppose that we are given the LLM model denoted as $f$ described by its parameters $\theta$, hence representing a function mapping the input to the corresponding prediction, as described below:
\begin{equation}
    \begin{aligned}
        f_\theta(x)
         =\prod_{i=1}^{|y|} P_\theta\left(y_i \mid y_{<i}, x\right),
    \end{aligned}
\end{equation}
where $P_\theta$ is the probability function for generating the next token in the given sequence  y$ = (y_1, \ldots, y_{\lvert y\rvert})$, and $y_{<i} = \{ y_1, \dots, y_{i-1} \}$. Given an unlearned descriptor $\left(x_u, y_u\right)$ related to an unlearning instance $\mathcal{I}$ (e.g., public figures or copyright-protected information). Current approaches often indiscriminately update $\theta$ to $\theta^{\prime}$ to ensure that all responses, $y_u^{\prime}=f_{\theta^{\prime}}\left(x_u\right)$, related to $\mathcal{I}$ are non-harmful. Yet, not all knowledge tied to $\mathcal{I}$ necessarily is required to be forgotten in this process.


In multi-modal unlearning, compared to unimodal unlearning, each sample contains multiple modalities, e.g., text and image pairs, and at least two modalities are processed for the MMU. Let:
\begin{equation}
D \;=\; \bigl\{\, \bigl(x_i^\text{(1)}, \dots, x_i^\text{(m)}, y_i \bigr)\bigr\}_{i=1}^{N},   
\end{equation}
where $m\ge 2$ is the number of modalities. We similarly define the \emph{forget set} $D_F$ of multi-modal samples that we aim to unlearn. The model $f_\theta$ is now capable of taking multi-modal input (e.g., a textual prompt plus an image) and producing an output $y$. The same \emph{forget} and \emph{retain} goals hold. Note that multimodal data allows unimodal unlearning (for example, one can mask tokens of omitted modalities), but multimodal unlearning requires processing of several modalities. From the practical point of view, both MU and MMU depend on optimization loss functions, which at the same time depend on the inputs, either unimodal or multimodal. Some methods, originally proposed for the MU, can naturally be applied for the MMU; however, some MU methods, for example RMU or $\text{SCRUB}_{\text{bio}}$ are non-trivial to be adapted for the MMU.


\subsection{Methods}
We briefly describe 5 top-performing MU methods from Tab. \ref{tab:regular_results} among all approaches described in detail in Appx.~\ref{sec:unlearning_methods}.

\paragraph{Retain Finetune} is a straightforward approach which involves finetuning the model on the retain set, assuming it will forget the knowledge from the $D_F$ while maintaining performance on the $D_R$. However, it is suboptimal for models with extensive pretraining, such as most LLMs.

\paragraph{IDK tuning} replaces original labels in the forget set $D_F$ with ''I don’t know'' responses while minimizing the loss on the retain set $D_R$ \cite{tofu2024}. 
The objective $L_{idk}$ ensures that the model retains performance on $D_R$ while aligning predictions on $D_F$ with uncertainty-based responses:
    \begin{equation*}
       L_{idk} = L(D_R, \theta) + L(D_F^{idk}, \theta)
    \end{equation*}

\paragraph{LLMU}was introduced in early LLM unlearning research \cite{yao2024largelanguagemodelunlearning}, optimizes the loss:  
\begin{multline*}
L_{\text{LLMU}} = -L(D_F, \theta) + L(D_F^{\text{idk}}, \theta)\\
+ \sum_{x,y \in D_R} \operatorname{KL}(p_{\theta}(y|x) || p_{\hat \theta}(y|x))   
\end{multline*}
Here, $\theta$ and $\hat \theta$ are model's parameters before and after unlearning, the first term promotes unlearning by maximizing loss on $D_F$, while the second reinforces forgetting using ''I don’t know'' labels instead of original targets. The KL-divergence term preserves performance on the retain set $D_R$ by aligning model outputs before and after unlearning. 



\paragraph{SCRUB (Teacher-Student)} formulates unlearning as a teacher-student setup, where a student model learns from a fixed teacher \cite{kurmanji2023unboundedmachineunlearning}. The student is optimized to match the teacher on the retain set $D_R$ while deviating on the forget set $D_F$. The loss function combines KL-divergence for retention ($L_R$), enforced divergence for unlearning ($L_F$), and task loss ($L_{\text{task}}$):
\begin{equation*}
    d(x, w^s) = \operatorname{KL}(p(f(x;w^o))||p(f(x;w^s))),
\end{equation*}

\begin{equation*}
    L_R = \frac{\alpha}{|D_R|} \sum_{x_r \in D_R} d(x_r, w^s),
\end{equation*}

\begin{equation*}
    L_F = \frac{1}{|D_F|} \sum_{x_f \in D_F} d(x_f, w^s),
\end{equation*}

\begin{equation*}
    L_{\text{task}} = \frac{\gamma}{|D_R|} \sum_{x_r \in D_R} l(x_r, y_r),
\end{equation*}

\begin{equation*}
    L_{\text{SCRUB}} = L_R - L_F + L_{\text{task}}
\end{equation*}
where $f(x;w^o)$ is the original teacher model with weights $w^o$, which are kept unchanged, $f(x;w^s)$ is the unlearned student model with parameters $w^s$, which are optimized.

\paragraph{Preference Optimization (DPO)} method applies Direct Preference Optimization (DPO) \cite{rafailov2023directpreferenceoptimizationlanguage} for MU and MMMU. The model is trained to reduce reliance on undesired information through a loss function combining task performance retention ($L_{\text{task}}$) and a DPO-based loss ($L_{\text{DPO}}$), which penalizes deviations from a reference model $\pi_{ref}$ fine-tuned on $D_F^{\text{idk}}$ with ``I don’t know'' labels as in IDK-tuning.





To sum up, we chose recently published MU methods for their easy adaptation to new modalities, needing only input data changes (text, images, or both) while maintaining core functionality. 

\section{\dataset{}}

\begin{figure}[t!]
\centering
\includegraphics[width=.15\textwidth]{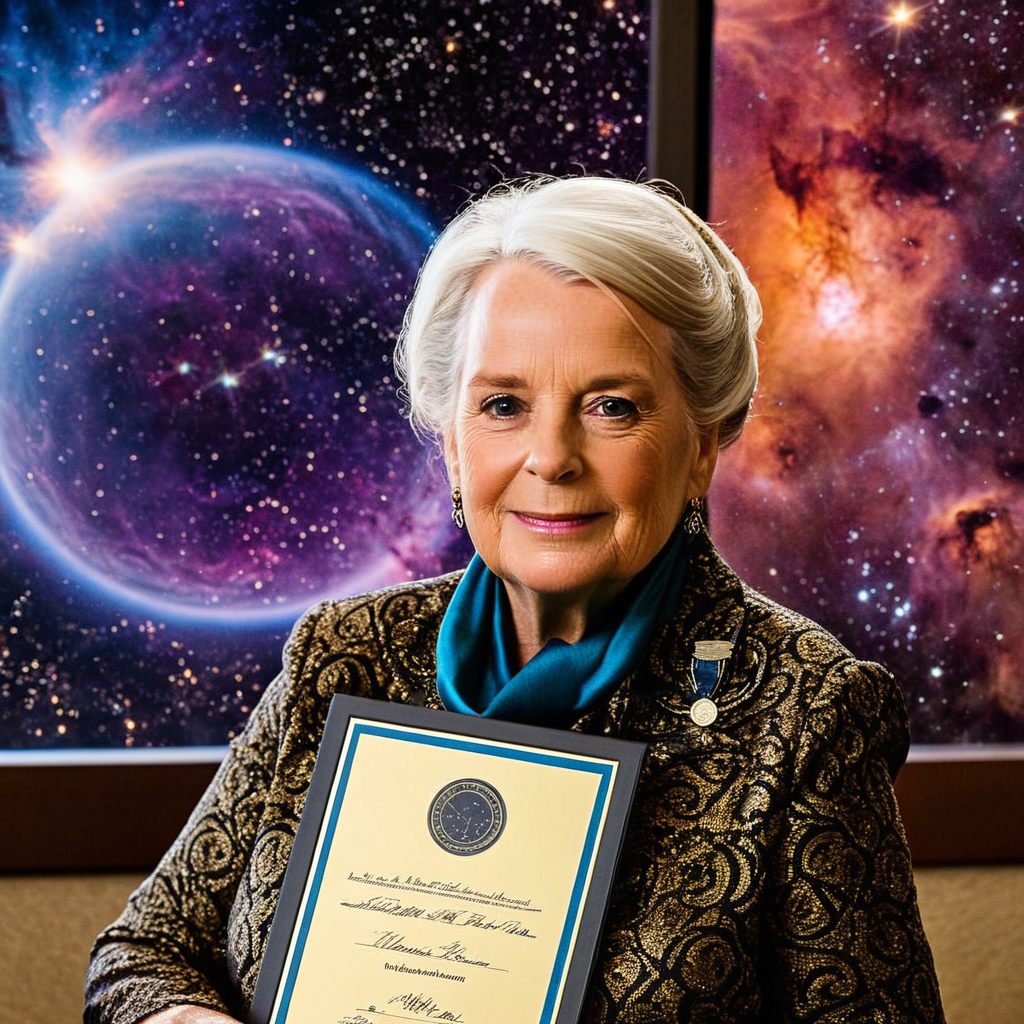}
\includegraphics[width=.15\textwidth]{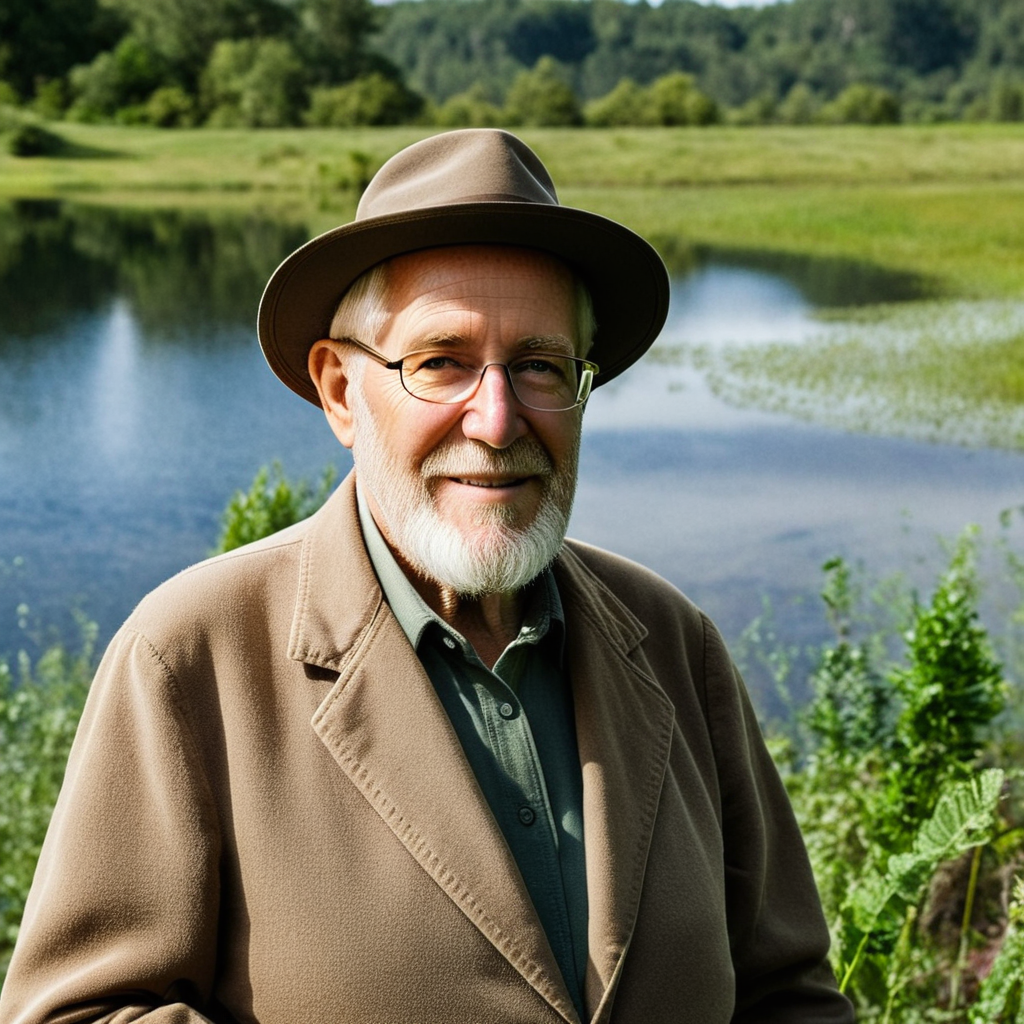}
\includegraphics[width=.15\textwidth]{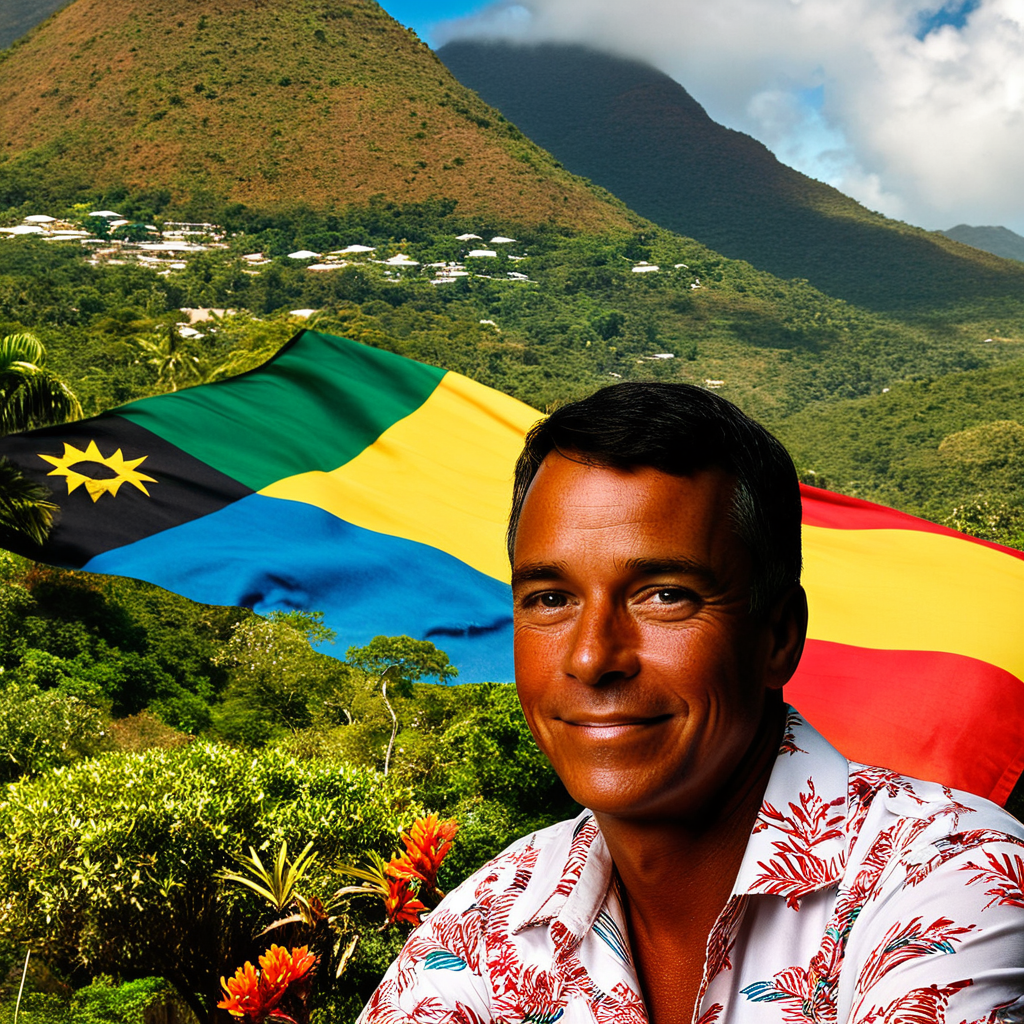} \\
\includegraphics[width=.15\textwidth]{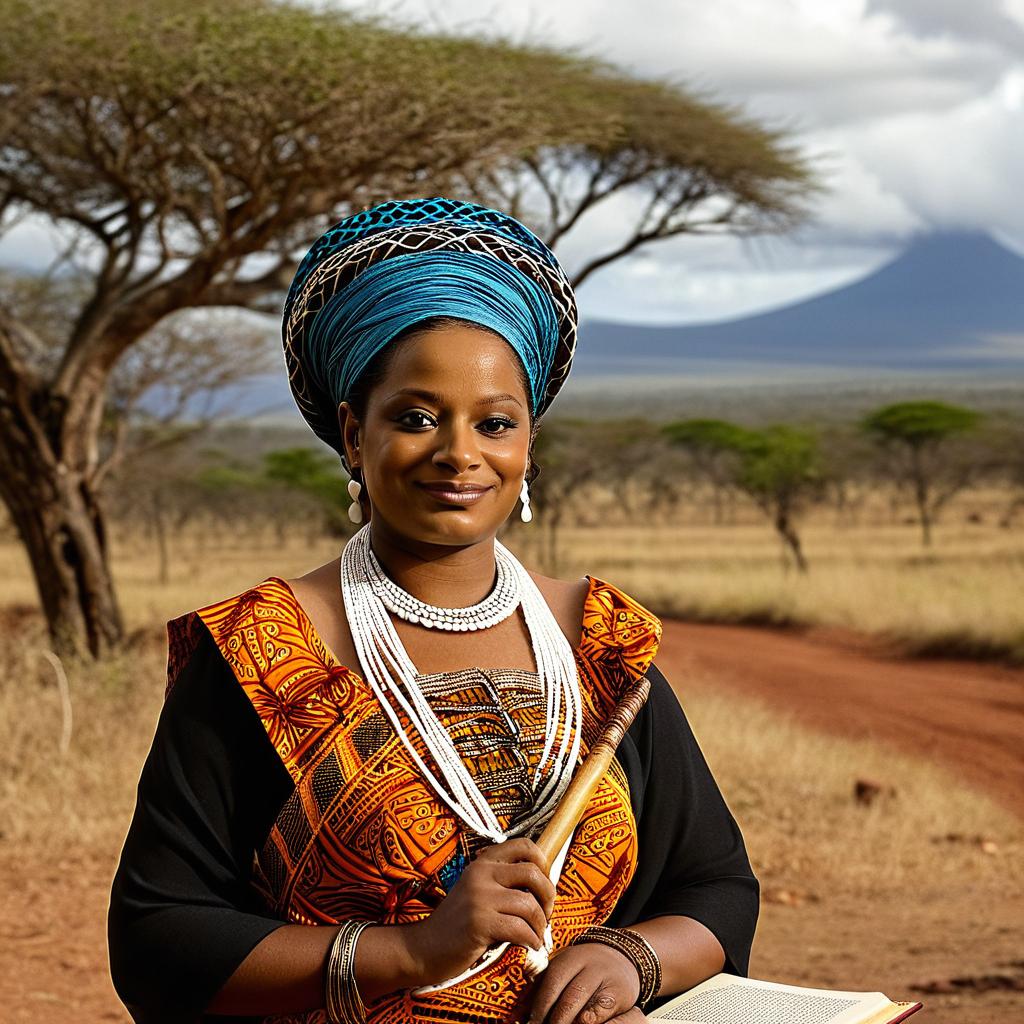} 
\includegraphics[width=.15\textwidth]{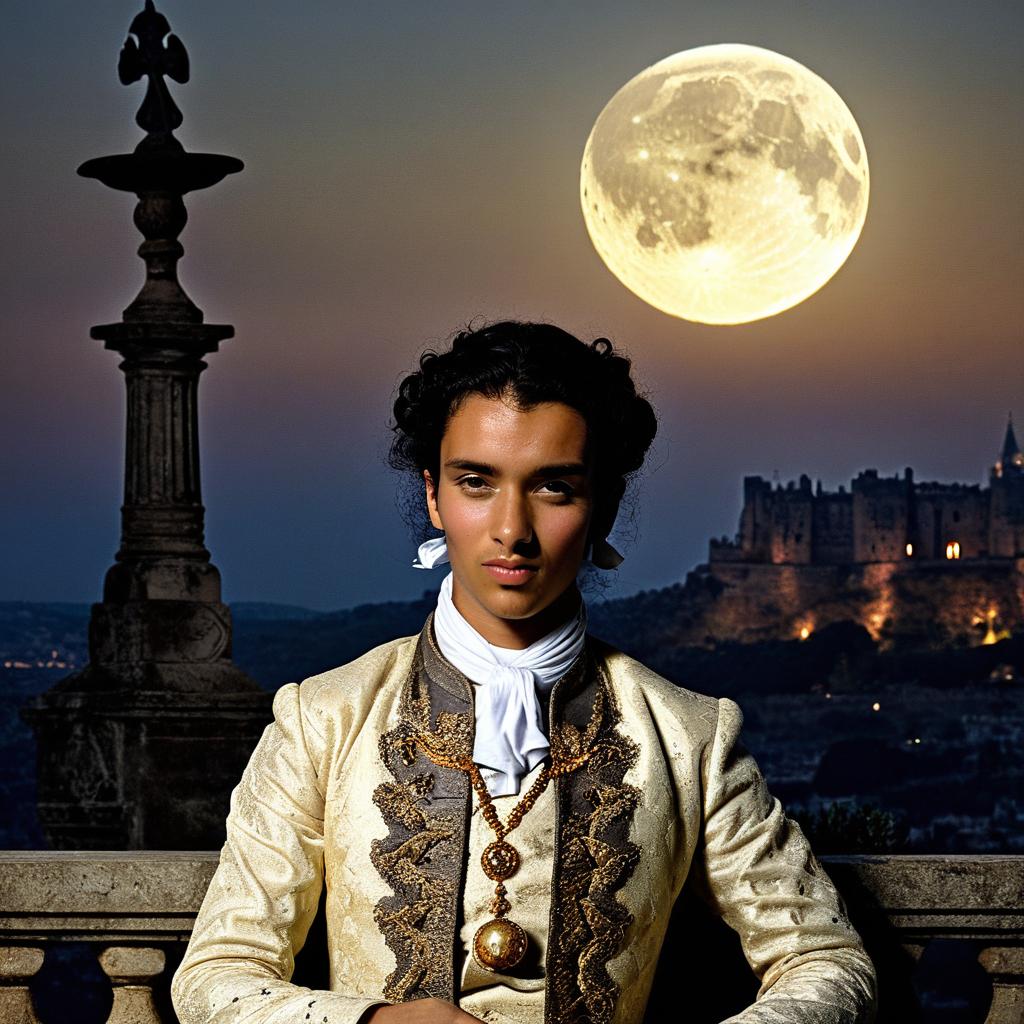}
\includegraphics[width=.15\textwidth]{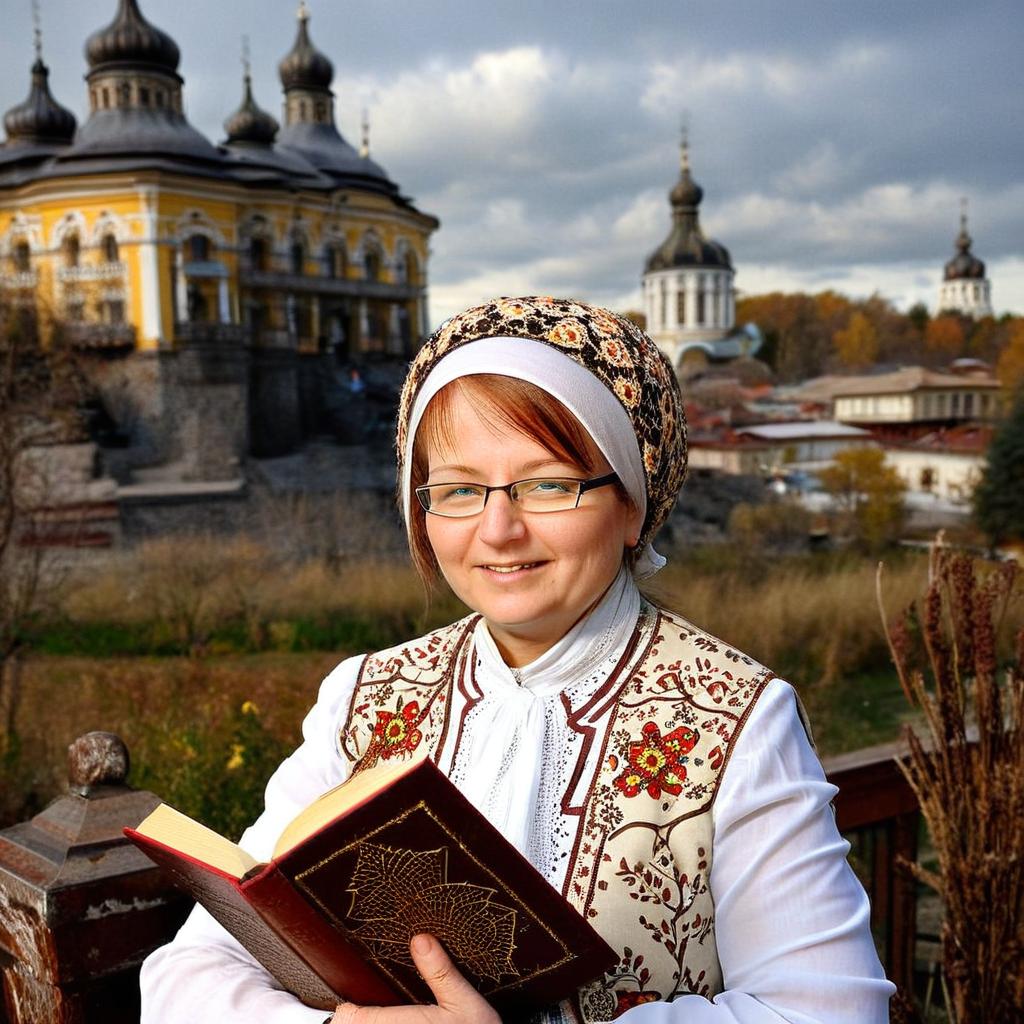} \\

\includegraphics[width=.15\textwidth]{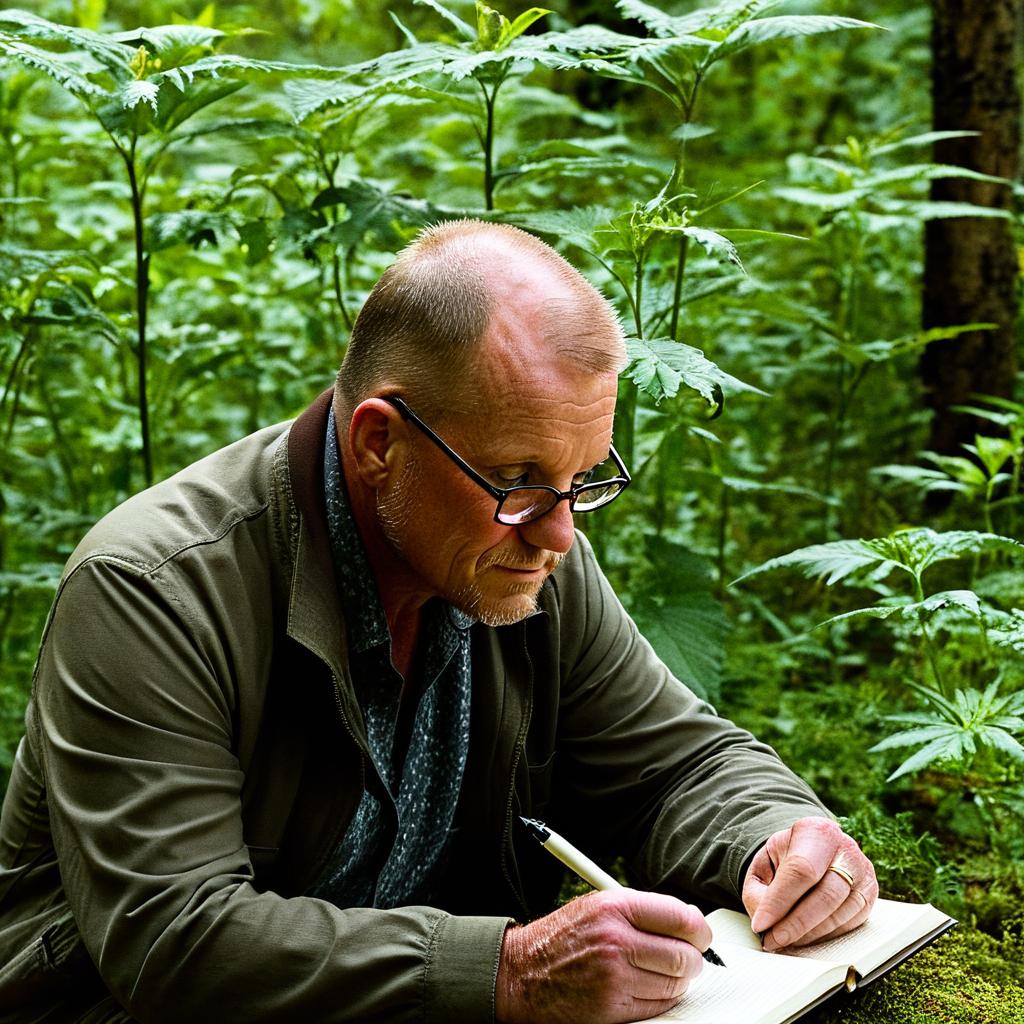} 
\includegraphics[width=.15\textwidth]{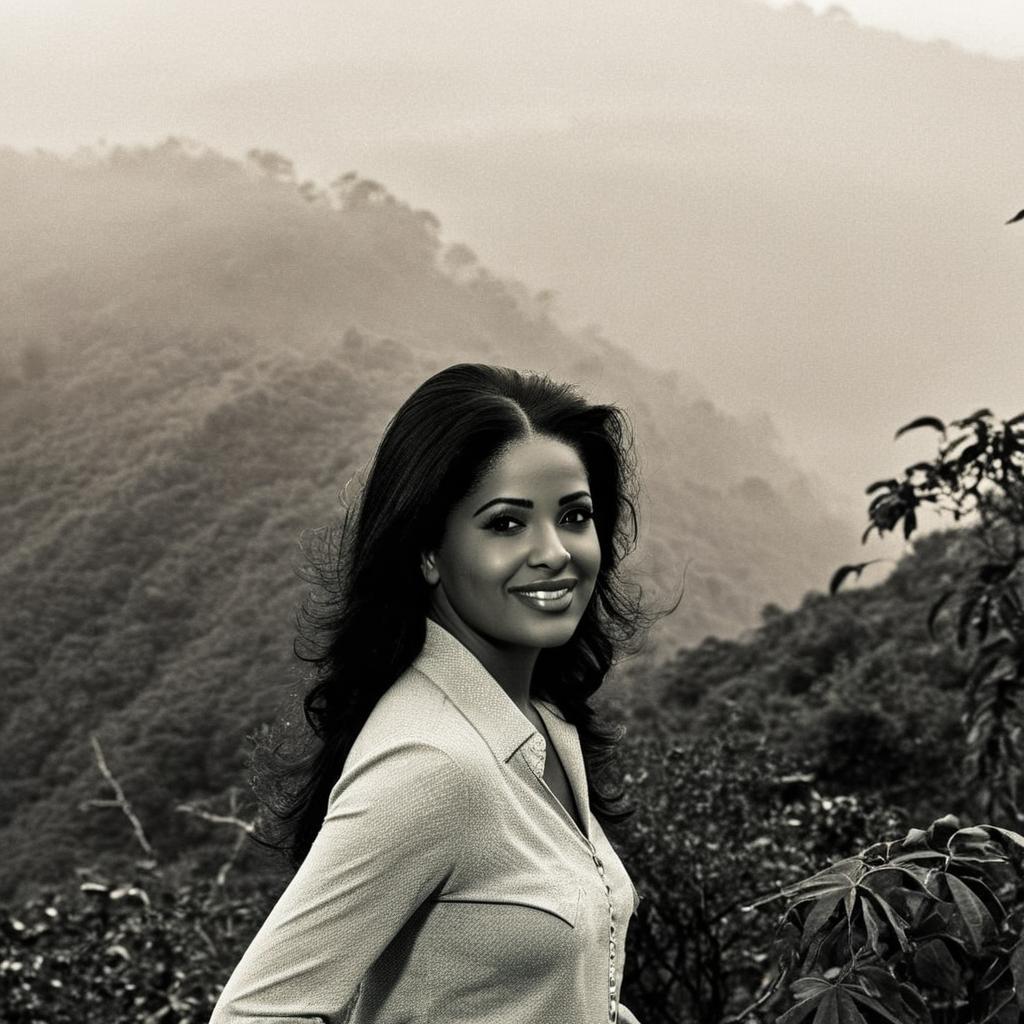}
\includegraphics[width=.15\textwidth]{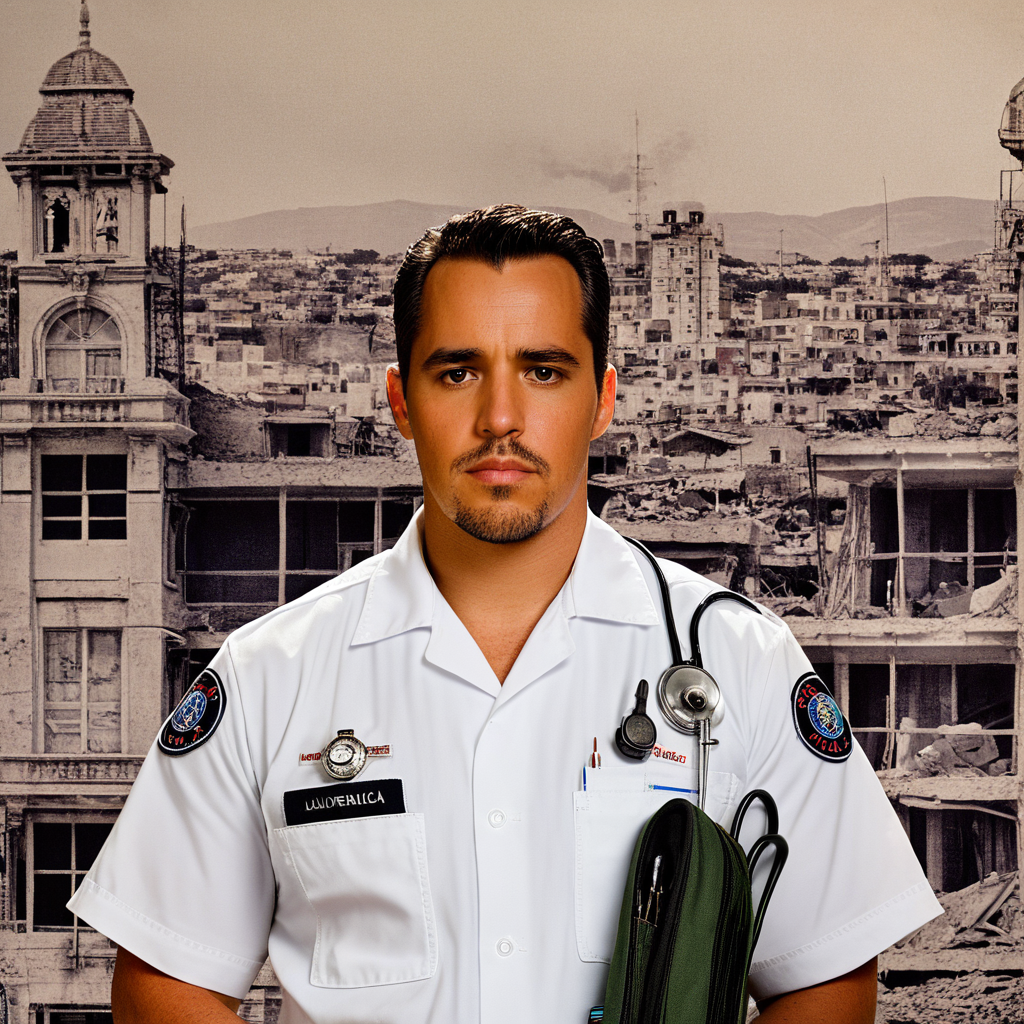} \\

\includegraphics[width=.15\textwidth]{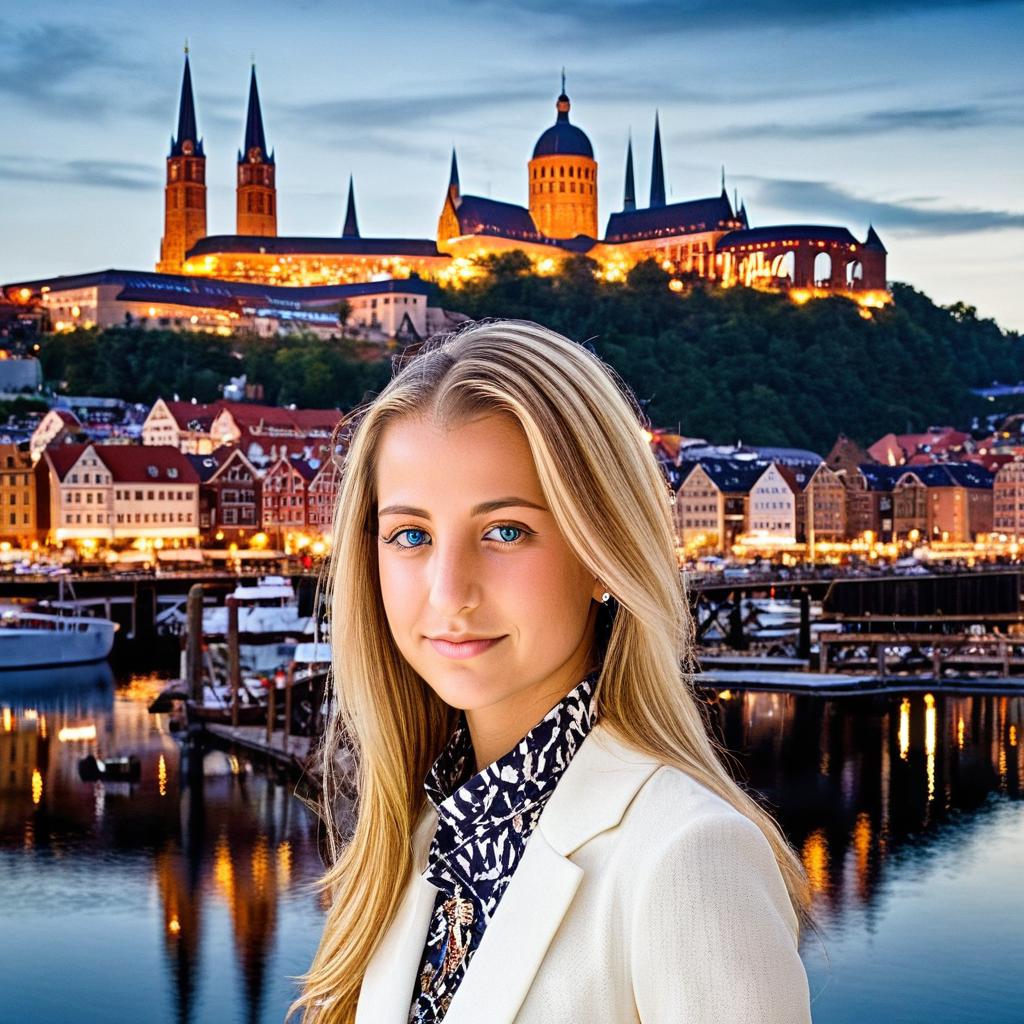} 
\includegraphics[width=.15\textwidth]{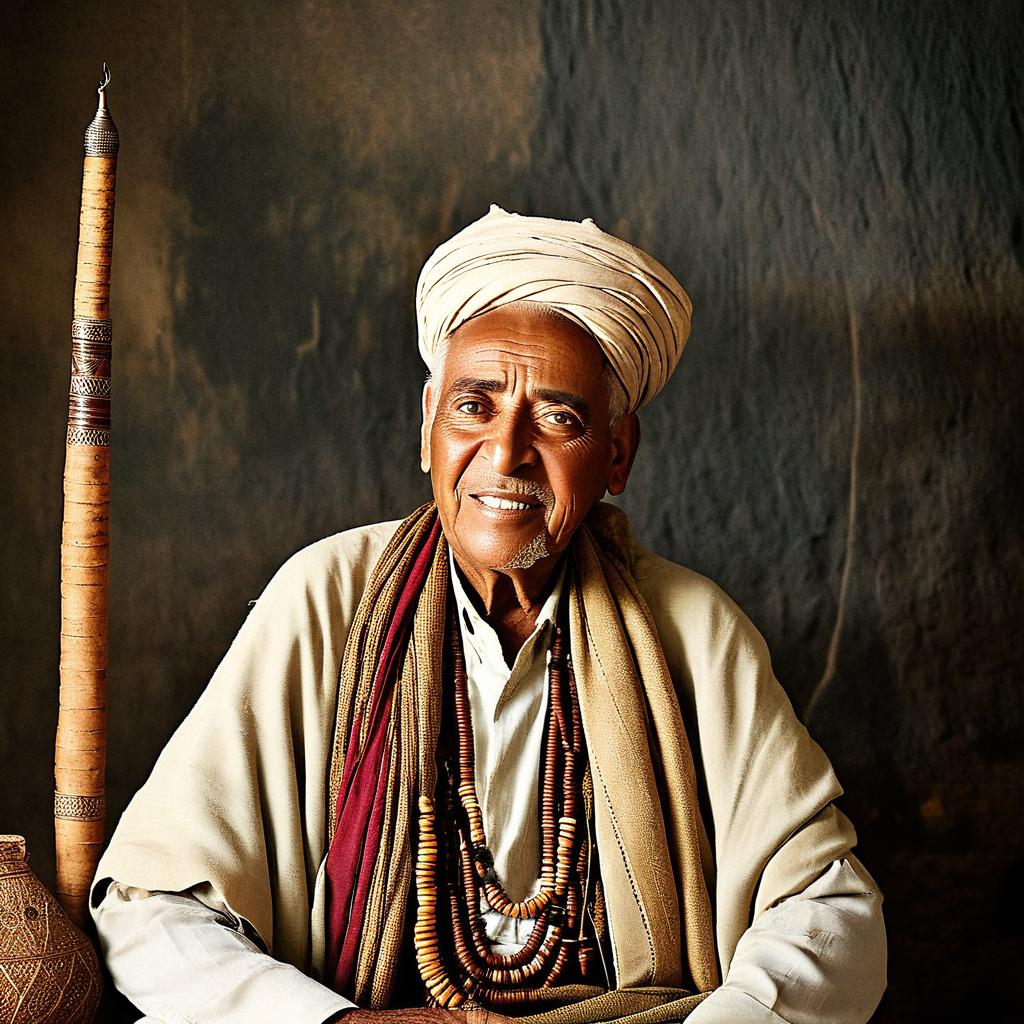}
\includegraphics[width=.15\textwidth]{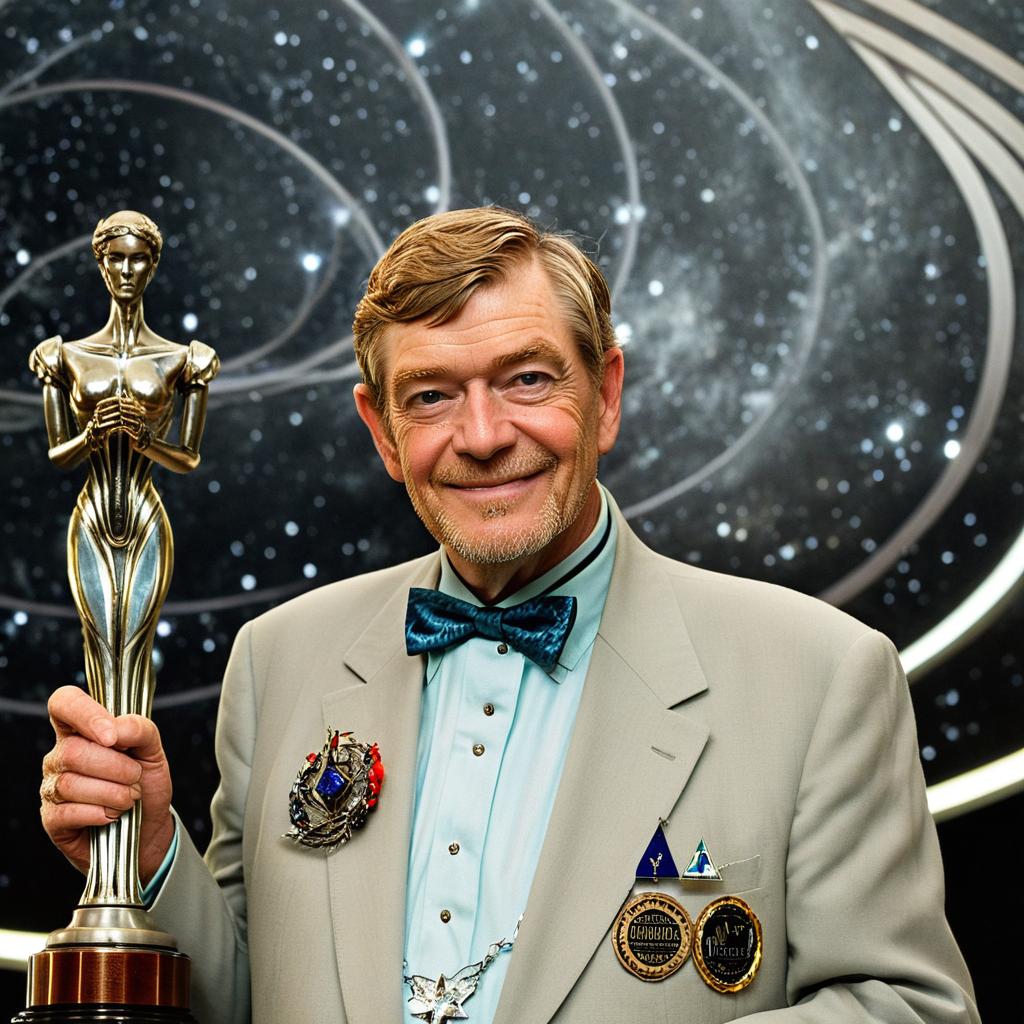} 
\caption{Examples of generated images showcasing a distinct individual from our dataset.}\label{fig:figure3}
\end{figure}

In this section, we describe a new benchmark \dataset{} designed for character unlearning. As a basis, our dataset utilizes the \textit{text-only} TOFU dataset \cite{tofu2024} within the same experimental setup to replicate a real-world scenario where privacy concerns arise in sensitive contexts. While external information from books, games, or movies is general knowledge to unlearn \cite{eldan2023whosharrypotterapproximate,li2024single, Xing2024EFUFEF}, character unlearning deals with specific contextual data that directly impacts individuals. This task addresses removing personal or confidential details and enhancing user privacy.


\subsection{Dataset Generation Process}
\begin{figure}[t!]
    \centering
    \includegraphics[width=\linewidth]{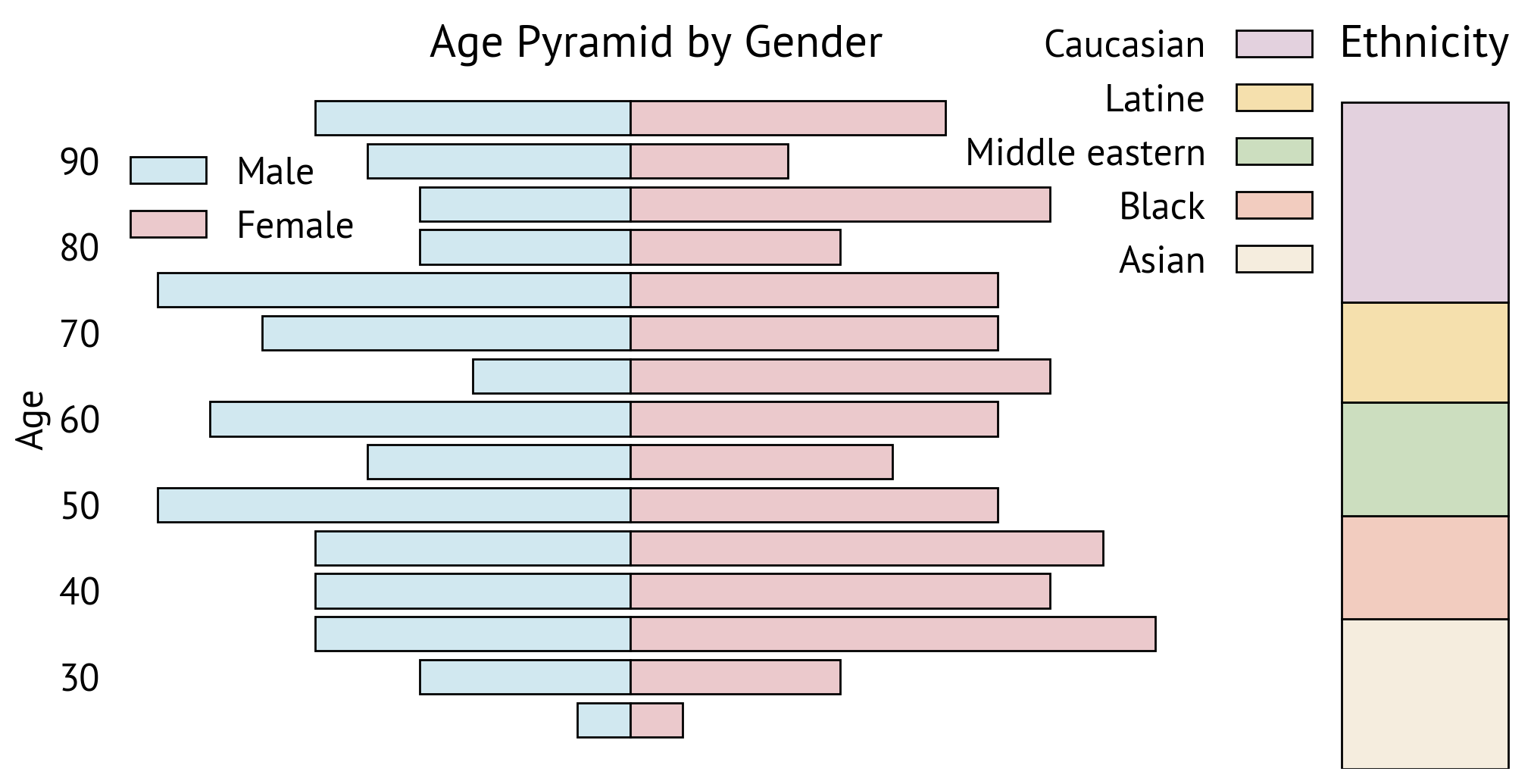}
\caption{Distributions of the attributes of the author's faces. We show that \dataset{} is balanced and representative regarding age, gender, and ethnicity.}
    \label{fig:face_age_distribution}
\end{figure}

The generation of synthetic faces for author profiles in our benchmark is motivated by ethical, technical, and practical considerations (see the complete rationale in Appx. \ref{sec:rationale}).
Firstly, for each of the 200 authors from the TOFU dataset, we extract their name, age, and ethnicity based on the knowledge provided in the original dataset. Also, we generate a pool of 2000 faces using StyleGAN2 \cite{karras2020analyzingimprovingimagequality} - an established generative model for face synthesis.
Each face is scored with a pre-trained image model to determine age, gender, and ethnicity. Then, for each author, we filter a pool of faces with similar characteristics and select the most appropriate one. We found out from textual information that the age distribution of the authors was highly shifted towards the older age group, so we needed to eliminate the age gap between authors' profiles and corresponding images. To do this, we used the image editing framework proposed in \cite{Bobkov_2024_CVPR} to shift the visual attributes of the faces to make them older. The final distribution of face and author characteristics is shown in Fig. ~\ref{fig:face_age_distribution}. After matching each author with a face, we used the diffusion model \cite{photomaker} to synthesize images based on the given face and corresponding prompt (Appx. ~\ref{sec:face_generation}).

\textbf{We perform a simple reality check to ensure the quality of generated faces.} We use the CLIP ViT-L/14 model, which is usually considered a visual encoder for most VLMs, to get embeddings of these three image sets -- our faces, CelebA \cite{liu2015faceattributes} and WebFace \cite{yi2014learning}. Then, we calculate the pairwise FID scores on top of the embeddings of these sets, and we get the following results: FID between our faces and CelebA is 74.4, between our faces and WebFace is 69.2, and between CelebA and WebFace is 62.1. This shows that the distance between our faces and the real-world faces is comparable to the distance between two real-world face datasets.
In addition to the author's face, the diffusion model needs a textual prompt to produce an image. We ask GPT-4 to generate these prompts from a question-answer pair from TOFU about an author. 
We generate $8$ images for each prompt, evaluate them using an ensemble of fake detection models, and select the most realistic one. Additionally, GPT-4o generates captions for each (image, visual prompt) pair, which are then included in the dataset to form pairs (image, caption). However, due to restrictions caused by GPT guard breaks and the identification of several bugs in the TOFU dataset (such as a nameless author), the final dataset includes fewer images than text pairs (3,770 compared to 4,000). 



\subsection{Splits}

We utilize four splits (sets) to evaluate MU (see Fig.~\ref{fig:four_datasets} for a sample from these splits): 
\\
\textbf{Forget.} Following methodology from \cite{tofu2024}, $D_F$ is made from data of 2, 10, and 20 persons  (1\%, 5\% and 10\%, respectively) of the full set $D$, consisting of 200 authors. This $D_F$ is expected to be unlearned by the model. 
\\
\textbf{Retain.} The retain set \(D_R\) consists of all data from the complete set \(D\) that is not in \(D_F\). The model should continue to work well on this subset and preserve its performance as much as possible. 
\\
\textbf{Real Faces.} To ensure the model retains knowledge of related concepts, such as faces, which are not present in the finetuning dataset, we evaluate it using a set of real-world faces. Specifically, we use the MillionCelebs dataset \cite{zhang2020global}, which consists of celebrity face-name pairs. We intersect this dataset with the most recognized celebrities from any year on the Forbes Celebrity 100 list to increase the likelihood that the model has seen these faces during pre-training. This results in a final set of 150 face-name pairs.
\\
\textbf{Real World.} To ensure that the model's overall visual capabilities remain intact throughout the unlearning process, we evaluate its performance on the Visual Question Answering (VQA) task using samples from \cite{grok15}.

\section{Experimental Setup and Evaluation}
\label{sec:metrics}
%

In this section, we briefly discuss the evaluation metrics and implementation details. 

\subsection{Evaluation Metrics}

We conduct a comprehensive evaluation using ROUGE-L, Probability Score, Truth Ratio, and Forget Quality metrics to thoroughly assess unlearning performance across textual, visual, and multimodal domains. Following \cite{tofu2024, li2024single,Xing2024EFUFEF}, this evaluation setup ensures that we capture the effectiveness of unlearning algorithms while examining both the retention and forgetfulness of information within the models. 

\paragraph{ROUGE-L.} 
ROUGE evaluates the \textit{word-level correspondence between the model's output and the ground truth answer} to a question. We calculate the ROUGE-L recall score \cite{lin2004rouge} by comparing the model's decoded output $f_{\hat \theta}(x)$ with the ground truth answer $y$ of the gold model $g$: $\text{ROUGE}(f_{\hat \theta}(x), y)$. This metric measures \textit{the model's remembrance of the knowledge in its exact formulations.}

\textbf{Probability Score.} 
One way to expose implicit knowledge from a model is through its logits, which are assigned to some factual tokens.
This metric assesses \textit{the model's capability to generate the correct answer}. 
We define the conditional probability $p(y|x)^{\frac{1}{|y|}}$ for input $x$ and answer $y$ (power $\frac{1}{|y|}$ corresponds to normalizing for length). Each input question $x$ is considered as a multiple choice question with possible answers $y_1, ..., y_n$, and then, assuming that $y_1$ is the correct answer, the desired probability score is computed as ${p(y_1|x)}/\left(\sum\limits_{i=1}^n p(y_i|x)\right)$. Higher values indicate better performance, revealing how well the model retains the correct answer.

\textbf{Truth Ratio} quantifies \textit{the alignment between predictions and the ground truth} by comparing the probability of a paraphrased correct answer against the averaged probabilities of several similarly formatted incorrect answers, providing insight into the effectiveness of the unlearning algorithm in removing specific information while maintaining overall accuracy. As defined by \cite{tofu2024}, assume that $\hat{y}$ denotes a paraphrased version of the answer $y$ for the input $x$ and $Y'$ is the set of 5 perturbations of the answer $y$. Then desired truth ratio $R$ is calculated as: $R=\frac{1}{\left|Y^{\prime}\right|}\left(\sum_{y^{\prime} \in Y^{\prime}} p\left(y^{\prime} \mid x\right)^{1 /\left|y^{\prime}\right|}\right) / p(\hat{y}\mid x)^{1 /|\hat{y}|}$.
This ratio is normalized and rescaled between 0 and 1, with higher values indicating better knowledge retention.

\textbf{Aggregate metrics.}
All three above-defined metrics are bounded between 0 and 1, so we combine them into a single metric to evaluate the overall performance. We set up the Real, Retain, and Forget metrics as a harmonic mean of the ROUGE, the Probability score, and the Truth Ratio computed on corresponding dataset splits.


\textbf{Forget Quality} calculates the ``distance'' of the unlearned model to the gold model, which is a proxy metric for the quality of \textit{exact} unlearning. Following \cite{tofu2024}, we take the Truth Ratios distribution of both models. However, instead of the p-value of the Kolmogorov-Smirnov test, we calculate the Jensen-Shannonn distance between these distributions. The latter metric better captures the differences between models, which we additionally check and describe in Appx. \ref{sec:forget_qual}. To maintain the higher - the better convention, we subtract the distance from 1.


\subsection{Implementation}
For the source model, we use LLaVa model \cite{liu2023llava} with ViT \cite{dosovitskiy2021imageworth16x16words} as visual encoder and LLaMa2-7B \cite{touvron2023llama2openfoundation} as a language model. First, we finetune it on the image captioning task using the full \dataset{}, both visual and textual parts. We call this model ``base'', as it contains forget and retain sets.
Then, we perform the unlearning process on it. We use the same hyperparameters for each method. We evaluate the unlearned model based on our metrics from Sec. \ref{sec:metrics}, using the Multi-choice VQA task for the probability score and the image captioning task for the Truth Ratio. For comparison, we also present the metrics of the ``gold'' model. Experimental results and metrics are shown in Tab. \ref{tab:regular_results}, with details provided in Appx. \ref{sec:multimodal_unlearning}. In addition, we perform experiments with the QwenVL2 model (Appx. \ref{sec:qwen_results}).

\section{Results}\label{s:results}

In the following sections, we describe the results of MU methods on our dataset. We seek to answer the following research questions:
\begin{enumerate}[label={\bfseries RQ\arabic*:}]
\item Does an unlearning method's performance on a single domain transfer directly to the performance in a multimodal setting? Should we study multimodal unlearning at all if we can easily predict its performance from single-domain experiments? 
\item Is unlearning only one modality (textual or visual) enough in a multimodal setup? How does the effectiveness vary depending on modality? How do different unlearning methods compare in their effectiveness for unlearning specific modalities?
\item In the context of multimodal unlearning, what methods perform the best?
\end{enumerate}

\subsection{Transferability from Single Domain}
To investigate \textbf{RQ1}, we analyze how well the performance of unlearning methods in single modalities predicts their effectiveness in multimodal settings. For the textual domain, we use the TOFU benchmark; for the visual, we use a standard U-MIA approach to the data, consisting of the faces from our dataset; the details for the pipelines and full results are provided in Appx. \ref{sec:textual_exps} and  \ref{sec:visual_exps}. For multimodal unlearning, we use our benchmark.

\textbf{The correlations between single-domain and multimodal (MM) rankings are relatively weak}. We rank methods according to their forget metric performance in each domain and calculate Spearman's rank correlation coefficient between single-domain and multimodal rankings. The correlation $\rho=0.7$ for text-MM and $\rho=0.2$ for visual-MM, indicating limited transferability (see Tab. \ref{tab:domain_comparison}).

\textbf{We observe significant discrepancies between single-domain and multimodal performance}. For example, LLMU achieves a retain metric of 0.51 in multimodal but degrades to 0.03 in the textual setting while maintaining good forget scores (0.25 vs 0.01). Similar patterns emerge for other methods, suggesting that single-domain evaluation is insufficient.

\textbf{Methods that perform well in single domains can fail catastrophically in multimodal settings}. For instance, RMU achieves good forget-retain balance in text-only (0.26/0.59) but completely fails on the retain set in multimodal setup (0.00/0.00), highlighting the unique challenges of multimodal unlearning.

\begin{tcolorbox}
\textbf{Takeaway 1}: Single-domain performance is a poor predictor of multimodal unlearning success, with relatively low-rank correlations ($\rho = 0.7$ and $\rho = 0.2$) and distinct failure modes. This emphasizes the need for dedicated multimodal evaluation frameworks and potentially new methods designed specifically for multimodal MU.
\end{tcolorbox}

\begin{table}[t!]
    \centering
    \scalebox{0.68}{
    \begin{tabular}{rccc}
    \toprule
    \textbf{Method} & \textbf{Text-only} & \textbf{Visual-only} & \textbf{Multimodal} \\
    & (Forget/Retain) & (Forget/Retain) & (Forget/Retain) \\
    \midrule
    LLMU & 0.01/0.03 & 85.2/88.9 & 0.25/0.51 \\
    DPO & 0.42/0.26 & 50.2/81.4 & 0.22/0.48 \\
    SCRUB & 0.42/0.26 & 42.59/99.4 & 0.36/0.52 \\
    IDK & 0.24/0.26 & N/A & 0.33/0.51 \\
    RMU & 0.59/0.26 & 67.9/99.0 & 0.00/0.00 \\
    Retain FT & 0.42/0.26 & 100.0/100.0 & 0.37/0.51 \\
    \midrule
    \multicolumn{4}{l}{\textit{Performance correlation with multimodal:}} \\
    Spearman's $\rho$ & 0.705  & 0.205  & 1.00 \\
    \bottomrule
    \end{tabular}}
    \caption{Transferability analysis across domains. We report forget (F) and retain (R) metrics for each method. Lower F and higher R are better. N/A indicates the method was not applicable. Correlation shows Spearman's rank correlation between single-domain and multimodal performance, with p-values in parentheses.}
    \label{tab:domain_comparison}
\end{table}

\subsection{Impact of Modality Selection on Unlearning}
To explore RQ2, we examine how the choice of unlearning modality impacts performance. We conduct experiments with three variants for each method: text-only, visual-only, and both modalities. Results in Tab. \ref{tab:modalities_exps} show distinct patterns across methods.

\paragraph{Text-only Unlearning.} Text-only approaches show mixed results. While some methods, like NPO, achieve good retain metrics (0.51) with moderate forget scores (0.29), others struggle significantly. RMU and GD completely fail on retain (0.01 and 0.00). KL shows middling performance with retain at 0.32 and forget at 0.23. This inconsistency suggests that text-only unlearning may disrupt cross-modal representations unpredictably.

\paragraph{Visual-only Unlearning.} Visual-only unlearning often achieves a better balance. DPO shows promising results with a forget metric of 0.23 while maintaining a 0.49 retain score. LLMU and SCRUB demonstrate similar patterns (forget: 0.37, 0.39; retain: 0.50, 0.49, respectively). However, some methods like RMU, GD, GA, and KL completely fail on retain metrics (0.00), indicating visual-only approaches are not universally successful.

\paragraph{Multimodal Unlearning.} Joint modality unlearning shows the most promising results for several methods. IDK improves its forget metric from 0.39 (text) and 0.35 (visual) to 0.33 (both) while maintaining stable retain performance (0.46). NPO shows strong real-world performance (0.49) but struggles with retain metrics when using both modalities. SCRUB demonstrates remarkable consistency across configurations (retain: ~0.50, forget: ~0.37), suggesting some methods are more robust to modality selection.
\begin{table}[t!]
    \centering
        \scalebox{0.8}{
\begin{tabular}{l@{}rcp{1cm}l@{}l@{}p{1cm}l@{}p{1cm}l@{}p{1cm}}
    \toprule
\textbf{Method}  & \textbf{Modality} & \makecell[tl]{\textbf{Real} $\uparrow$ \\ \textbf{metric}} & \makecell[tl]{\textbf{Retain} $\uparrow$ \\ \textbf{metric}} & \makecell[tl]{\textbf{Forget} $\downarrow$ \\ \textbf{metric}} & \makecell[tl]{\textbf{Forget} $\uparrow$\\ \textbf{Quality} } \\
    \midrule
    Gold & --- & 0.50 & 0.51 & 0.19 & 1.00 \\
    Base & --- & 0.48 & 0.51 & 0.35 & 0.85 \\ \hline \hline
    RMU & text & 0.31 & 0.01 & 0.02 & 0.75 \\
    RMU & visual & 0.24 & 0.00 & 0.00 & 0.75 \\
    RMU & both & 0.22 & 0.00 & 0.00 & 0.80 \\ \hline
    GD & text & 0.26 & 0.00 & 0.00 & 0.79 \\
    GD & visual & 0.29 & 0.00 & 0.00 & 0.20 \\
    GD & both & 0.49 & 0.51 & 0.37 & 0.85 \\ \hline
    Retain FT & text & 0.49 & 0.51 & 0.37 & 0.85 \\
    Retain FT & visual & 0.46 & 0.45 & 0.42 & 0.85 \\
    Retain FT & both & 0.46 & 0.46 & 0.40 & 0.85 \\ \hline
    GA & text & 0.34 & 0.00 & 0.00 & 0.22 \\
    GA & visual & 0.29 & 0.00 & 0.00 & 0.32 \\
    GA & both & 0.29 & 0.00 & 0.00 & 0.32 \\ \hline
    KL & text & 0.49 & 0.32 & 0.23 & 0.71 \\
    KL & visual & 0.29 & 0.00 & 0.00 & 0.28 \\
    KL & both & 0.48 & 0.35 & 0.27 & 0.81 \\ \hline
    IDK & text & 0.48 & 0.50 & 0.39 & 0.85 \\
    IDK & visual & 0.44 & 0.45 & 0.35 & 0.84 \\
    IDK & both & 0.46 & 0.46 & 0.33 & 0.84 \\ \hline
    NPO & text & 0.51 & 0.51 & 0.29 & 0.85 \\
    NPO & visual & 0.48 & 0.43 & 0.24 & 0.84 \\
    NPO & both & 0.49 & 0.00 & 0.00 & 0.72 \\ \hline
    SCRUB & text & 0.49 & 0.51 & 0.37 & 0.85 \\
    SCRUB & visual & 0.48 & 0.49 & 0.39 & 0.85 \\
    SCRUB & both & 0.49 & 0.51 & 0.37 & 0.85 \\ \hline
    DPO & text & 0.47 & 0.45 & 0.42 & 0.85 \\
    DPO & visual & 0.48 & 0.49 & 0.23 & 0.84 \\
    DPO & both & 0.46 & 0.47 & 0.28 & 0.84 \\ \hline
    LLMU & text & 0.48 & 0.46 & 0.40 & 0.85 \\
    LLMU & visual & 0.49 & 0.50 & 0.37 & 0.85 \\
    LLMU & both & 0.47 & 0.48 & 0.33 & 0.84 \\
    \bottomrule
    \end{tabular}
    }
        \caption{Results of unlearning of different modalities within multimodal setup. We finetune on full datasets (both modalities), then forget on a single domain subset (text or visual) or full forget set. Base -- model before unlearning. Gold - a model trained only on retain. }
    \label{tab:modalities_exps}
\end{table}
\begin{tcolorbox}
\textbf{Takeaway 2}: While visual-only unlearning often outperforms text-only approaches, the effectiveness varies significantly by method. Methods like SCRUB maintain consistent performance across modalities (retain: 0.49-0.51), while others show dramatic variations. NPO and KL demonstrate that combining modalities can improve forget quality (0.72-0.81) compared to single-modality approaches (0.28-0.85). However, the optimal choice of modality depends heavily on the specific method and desired performance trade-offs.
\end{tcolorbox}

\subsection{Unlearning Both Domains}
\label{sec:unlearning_both}
Having established that multimodal unlearning requires addressing both modalities, we evaluate all available unlearning methods on our source model $f_{\theta}$ across both domains. For these experiments, we use a forget set containing data about 20 persons (10\% of the dataset), encompassing both their textual and visual information.

As shown in Tab. 3, there are three distinct categories of method behaviour. GA, GD, KL, and RMU achieve perfect unlearning (forget metric = 0) but completely destroy the model's retained knowledge (retain metric = 0). IDK, SCRUB, and Retain FT maintain strong retain performance (~0.51) but struggle with effective forgetting (forget metrics 0.33-0.37). LLMU and DPO balance forgetting and retention best, maintaining reasonable retain metrics (0.48-0.51) while showing improved forget performance (0.22-0.25).

\begin{tcolorbox}
    \textbf{Takeaway 3}: Most unlearning methods struggle with the trade-off between effective forgetting and knowledge retention in multimodal settings. Only LLMU and DPO show promise in balancing these objectives, but their performance remains below the gold model (forget = 0.19, retain = 0.51).
\end{tcolorbox}    

\section{Conclusion}

In this work, we introduce \dataset{}, the first open-sourced benchmark designed to assess machine unlearning in a textual-visual multimodal setup. Our evaluation of existing unlearning techniques across domains shows that multimodal unlearning is more challenging than previously anticipated, laying the ground for further research. Our findings offer a new perspective than earlier results on safety alignment \cite{chakraborty-etal-2024-textual}, which suggested that text-only unlearning is sufficient for multimodal models. 

While \dataset{}’s synthetic personas ensure controlled evaluation, real-world data (e.g., diverse facial features, noisy captions) may introduce new challenges. Additionally, our study focuses on visual-language models, leaving other modalities (e.g., audio, video) unexplored. By open-sourcing \dataset{} and establishing the first multimodal MU leaderboard, we aim to accelerate progress toward ethical, privacy-preserving multimodal AI. Our findings highlight that MMU is not merely an extension of unimodal unlearning but a distinct challenge requiring novel methodologies.




\section*{Limitations}

Despite the contributions of this work, several limitations remain that need further investigation. One major limitation is the reliance on synthetic data, as CLEAR is based on such dataset, which may not fully capture the complexity of real-world scenarios, thus limiting the generalizability of our findings. Additionally, while our work focuses on unlearning methods designed for privacy-centric applications, such as removing personal data, it may not fully address other unlearning needs, such as removing harmful content. Moreover, our benchmark mainly evaluates fine-tuning-based unlearning methods using sophisticated loss functions, leaving unexplored other broader unlearning techniques, such as analytical or mechanical approaches. Another challenge lies in the scalability of these unlearning methods, as they may struggle to scale efficiently when applied to larger models and datasets, hindering their potential use in real-world systems. Furthermore, our focus on catastrophic forgetting overlooks unintended side effects, such as the introduction of biases or the degradation of model performance on unrelated tasks, and the broader impact of unlearning on fairness and safety remains an open area for future research.

\section*{Ethics}

In this work, we focus on \textit{unlearning character-specific knowledge} in pre-trained visual-language models (VLMs). We aim to enable VLMs to selectively forget all traces of specific synthetic personas—including their textual biographies, visual appearances, and cross-modal associations—while preserving the model’s general capabilities. This addresses critical ethical concerns, such as the \textit{right to be forgotten} and prevention of unintended memorization. For forget and retain sets, all data is synthetically generated to avoid biases and leakage from real-world sources, with evaluation protocols encouraging responsible use. These sets were manually checked by one of the authors. Datasets on celebrity recognition and general VQA are publicly accessible sources. We also urge researchers and developers to employ our methods responsibly and with ethical considerations.

We used 84 hours of A100 GPU computation, resulting in an estimated 9 kg of CO2 emissions.

\paragraph{Use of AI Assistants.} 


We utilize Grammarly to enhance and proofread the text of this paper, correct grammatical, spelling, and stylistic errors, as well as rephrase sentences. Consequently, certain sections of our publication may be identified as AI-generated, AI-edited, or a combination of human and AI contributions.

\section*{Acknowledgements}
The work was supported by a Research Center for Trusted Artificial Intelligence the Ivannikov Institute for System Programming of the Russian Academy of Sciences.


\appendix

\include{appendix}

\end{document}

%% file: appendix.tex
\section{Rationale for Synthetic Face Selection} 
\label{sec:rationale}
The use of synthetic faces, rather than real-world facial data, in our benchmark is motivated by ethical, technical, and practical considerations. First, synthetic faces eliminate privacy risks and ethical concerns associated with real facial datasets. By generating artificial personas, we avoid biases inherent in real-world datasets and ensure no real individuals are misrepresented, aligning with the \textit{right to be forgotten} principle.   

Second, synthetic data provides precise control over memorization evaluation. Real faces risk contamination from prior model training (e.g., pre-existing celebrity images in model weights), which could confound unlearning performance measurements. Synthetic faces, being novel and never publicly released, guarantee that models learn exclusively from our benchmark, enabling accurate assessment of unlearning efficacy. Notably, our experiments reveal that models still struggle to fully erase synthetic faces -- despite their controlled generation. This implies that applying current unlearning methods to real-world faces (e.g., from public sources like Wikipedia) would face greater challenges, as real data introduces uncontrolled variability and pre-existing biases that synthetic benchmarks deliberately exclude. In other words, applying artificial profiles ensures that the considered model has not seen the authors during pre-training, and this is essential for the fair evaluation of MU methods,  as we can easily compare MU results with a \emph{gold} model, which has never seen the profiles we want to forget, without expensive re-training from scratch on a large plethora of data required for LLMs and VLLMs training. 

Third, synthetic generation prevents cross-modal leakage. By explicitly linking synthetic faces to their textual biographies, we isolate memorization tests to our dataset, ensuring no external knowledge interferes. This allows rigorous evaluation of whether unlearning a biography also removes its associated face.  

Additionally, synthetic faces enhance reproducibility and scalability. Unlike real datasets burdened by licensing restrictions, synthetic data can be freely shared, fostering open benchmarking. On-demand generation also supports customizable testing, such as expanding the forget set to thousands of unique identities without legal barriers. Our comprehensive image generation strategy suits the author's textual descriptions and preserves consistency among his or her images. Still, it enables sufficient diversity between different authors regarding age, gender and ethnicity.

Also, it is worth noting that synthetic data is also used in practice in literature for unlearning-related task. In \cite{zhang2024unlearncanvasstylizedimagedataset}, the authors introduce a novel dataset, UnlearnCanvas, designed to benchmark machine unlearning techniques in diffusion models, offering a comprehensive, high-resolution stylized image dataset to evaluate the unlearning of artistic styles and associated objects. The UnlearnCanvas dataset includes generated images across 60 artistic painting styles, with 400 images per style across 20 object categories. The dataset facilitates the quantitative evaluation of vision generative modelling tasks, including machine unlearning, style transfer, vision in-context learning, bias removal for generative models, and out-of-distribution learning. The paper \cite{ma2024benchmarkingvisionlanguagemodel} introduces a new benchmark, FIUBench, to evaluate the effectiveness of unlearning algorithms in Vision Language Models under the Right to be Forgotten setting. The authors formalize the VLM unlearning task and construct a Fictitious Facial Identity VQA dataset of synthetic faces paired with randomly generated personal information to study privacy under the Right to be Forgotten scenario. This approach allows precise control over the source of information and its exposure in the unlearning dataset. The dataset includes personal backgrounds, health records, and criminal histories for each facial identity. The work \cite{dhasade2024quickdropefficientfederatedunlearning} introduces a novel approach to Federated Unlearning, which aims to effectively remove specific training data knowledge from machine learning models trained through Federated Learning. The authors highlight the inefficiencies of existing Federated Unlearning methods that often involve high computational costs due to gradient recomputation and storage requirements. The provided approach, QuickDrop, is designed to streamline the unlearning process by generating compact synthetic datasets that represent the gradient information used during model training. This approach significantly reduces the volume of data needed for unlearning while maintaining performance efficiency. QuickDrop employs a method called dataset distillation to create a compact dataset that captures essential features of the original training data. This dataset is approximately 1\% of the size of the original data, leading to minimal storage overhead. Each client generates a synthetic dataset through gradient matching, which serves as a compressed representation of their original gradients.

In summary, synthetic faces prioritize ethical rigour, experimental precision, and reproducibility—critical for advancing multimodal machine unlearning research. The observed difficulty in unlearning even synthetic faces underscores fundamental model limitations, which real-world deployments (e.g., authors' faces) would exacerbate due to added complexity. Our benchmark thus serves as a necessary precursor to addressing practical challenges in ethical AI.

\section{Unlearning Methods}
\label{sec:unlearning_methods}
This section describes the main unlearning approaches considered in this work.
\begin{enumerate}
  \item \textbf{Finetuning on retain data.}\label{finetuning} The most straightforward method to conduct unlearning is to finetune the model on the retain set, assuming that 
  the model will unlearn the knowledge from the forget set and preserve its performance on the retain set. Despite its simplicity and reasonable effectiveness for relatively small models, it is not usable in models with huge sizes of pre-train sets, such as most LLMs. 

  \item \textbf{Gradient ascent on forget set.} \label{grad_ascent} In this method, unlearning is done by maximizing the loss on forget data with the intuition that it will lead to getting predictions that are dissimilar from the correct answers for forget set and consequently unlearning desired information. Thus, this method can be considered as a finetuning procedure with the following loss function:
    \begin{equation*}
    L(D_F, \theta) = \frac{1}{|D_F|} \sum_{x \in D_F} {NLL}(x, \theta),
    \end{equation*}

  where $\operatorname{NLL}(x, \theta)$ is the negative log-likelihood of the model on the input $x$.

Instead of maximizing the $\operatorname{NLL}$ loss, maximizing the entropy 
  of the model's predictions on the forget set is possible. The intuition behind this trick is that it will correspond to the increase of the model's uncertainty in its predictions on forget set, which will also correspond to successful unlearning.

  \item \textbf{Gradient difference.} \label{grad_diff} \cite{liu2022continuallearningprivateunlearning}
  The next method builds on the concept of combining two previous methods. It aims to 
  increase the loss on the forget data and at least maintain the loss on the retain set. The 
  loss function is defined as follows:
  \begin{equation*}
  L_{GD} = - L(D_F, \theta) + L(D_R, \theta),
 \end{equation*}
  where $D_F$ is the forget set that remains constant, $D_R$ is the retain set that is randomly sampled during training, and $L$ is a suitable loss function.


  \item \textbf{KL minimization}. \label{KL} This approach aims to minimize the Kullback-Leibler ($\operatorname{KL}$) divergence between the model's predictions on the retain set before and after unlearning while maximizing the conventional loss on the forget set. 
The $L_{\text{KL}}$ loss function is defined as
\begin{equation*}
\begin{split}
\frac{1}{|D_F|} \sum_{x \in D_F} \frac{1}{|s|} \sum_{i=2}^{|s|} \operatorname{KL}\left( P(s_{<i}|\theta) \middle\| P(s_{<i}|\theta') \right).
\end{split}
\end{equation*}
The total objective function is formulated as follows:
\begin{equation*}
    L_{obj} = - L(D_F, \theta) + L_{\text{KL}},
\end{equation*}

  where $\theta'$ is the model's weights before unlearning, $s$ is the input sequence, $L$ is conventional loss, and $P(s| \theta)$ 
  is the model's logits on the input sequence $s$ with weights $\theta$.

  \item \textbf{IDK tuning}\label{idk}. Introduced in \cite{tofu2024}, this method aims to minimize the loss on the retain set,
  meanwhile, it uses pairs of inputs and "I don't know"(or some variations) labels instead of the original labels on the forget set. 
  The loss function is defined as follows:
    \begin{equation*}
       L_{idk} = L(D_R, \theta) + L(D_F^{idk}, \theta),
    \end{equation*}
  where $L$ is some loss function, $D_R$ is retain set, and $D_F^{idk}$ is forget set with labels replaced with "I don't know" answers or some variations of them.

  \item  \textbf{Preference Optimization}\label{DPO}. Inspired by Direct Preference Optimization (DPO) \cite{rafailov2023directpreferenceoptimizationlanguage}, the unlearning task can be framed as a preference optimization problem. In DPO, the model is trained to optimize user preferences directly, typically by maximizing the alignment between the model's outputs and the user's desired outcomes. Similarly, the goal of unlearning can be viewed as removing specific knowledge or patterns that the model has learned, effectively optimizing the model's outputs to align with new preferences that exclude the undesired information.

In this context, the unlearning task aims to adjust the model's parameters such that the output reflects a change in the learned distribution, making the model "forget" specific pieces of knowledge. This can be formalized as a preference optimization problem, where the preference is towards outputs that no longer rely on unwanted data. Let $L$ represent the loss function used for this task, which balances the model's performance on new data and its ability to unlearn specific information.

A common approach is to use a loss function that minimizes the difference between the model's current predictions and the desired "unlearned" predictions of the chosen reference model. The following loss function was considered to optimize for unlearning:


\begin{equation*}
L = \lambda_1 L_{\text{task}}(D_F^{idk}, \theta) + \lambda_2 L_{\text{DPO}}(\pi_{\theta}, \pi_{ref}),
\end{equation*}

\begin{equation*}
\begin{split}
 &L_{DPO}(\pi_{\theta}, \pi_{ref}) = \\ 
 &= -\mathop{\mathbb{E}_{x, y \in D_F}}_{y' \in D_F^{idk}} \Big[ 
    \log \sigma(\beta\log\frac{\pi_{\theta}(y'|x)}{\pi_{ref}(y'|x)} - \\ 
    &-\beta\log\frac{\pi_{\theta}(y|x)}{\pi_{ref}(y|x)}{}) \Big],
 \end{split}
 \end{equation*}
where $\pi_{\theta}$ is related to the unlearned model which we try to optimize, $\sigma$ is the sigmoid function, $\pi_{ref}$ is reference model which in our case is fine-tuned on $D_F^{idk}$ data, where labels are replaced with "I don’t know" answers, $(x, y)$ is input-answer pair from the forget set, $y'$ is "I don't know"-like answer corresponding to this pair, $L_{\text{task}}(D_F^{idk}, \theta)$ is the standard task loss (e.g., cross-entropy) on the set $D_F^{idk}$, and $L_{\text{DPO}}(\pi_{\theta}, \pi_{ref})$ is DPO loss used for unlearning, which penalizes the model for retaining unwanted knowledge, computed between the input data $x$ and the undesired in terms of unlearning labels $y$. $\lambda_1$ and $\lambda_2$ are weighting coefficients that balance the trade-off between task performance and the unlearning process (equal to 1 both), and $\beta$ is the DPO coefficient (taken as 0.1 in our setting).

This formulation allows the model to optimize for maintaining task performance while ensuring the forgetting of specified information, similar to the dual objective in preference optimization. In the same way that DPO tailors the model to user preferences, this method shapes the model to "prefer" forgetting certain information, effectively unlearning it.

  \item \textbf{Negative Preference Optimization} \label{NPO}. Proposed in \cite{zhang2024negativepreferenceoptimizationcatastrophic} this method
  can be treated as DPO without positive examples. In our setting, the final loss function  $L_{NPO}$ for this method is derived as follows:
\begin{equation*}
\frac{2}{\beta} \mathop{{}\mathbb{E}_{x, y \in D_F}} \Big[ 
    \log \left(1 + \Big(\frac{\pi_{\theta}(y|x)}{\pi_{ref}(y|x)} \Big)^{\beta}\right) \Big],
\end{equation*}where all the notation is the same as for the previous DPO method. $\beta$ was also taken equal to 1. Such loss functions ensure that the model output probability $\pi_{\theta}(y|x)$ is as small as possible, corresponding to the unlearning objective of the forget data.

  \item \textbf{Teacher-Student (SCRUB)} \label{scrub}\cite{kurmanji2023unboundedmachineunlearning} The main idea
  of this method is to train a student model, which is taken as a desired unlearned model from the original one, such 
  that it will ``disobey'' the teacher original model on the forget set.
  The resulting loss of student model in this method is constructed as follows:
\begin{equation*}
    d(x, w^s) = \operatorname{KL}(p(f(x;w^o))||p(f(x;w^s))),
\end{equation*}

\begin{equation*}
    L_R = \frac{\alpha}{|D_R|} \sum_{x_r \in D_R} d(x_r, w^s),
\end{equation*}

\begin{equation*}
    L_F = \frac{1}{|D_F|} \sum_{x_f \in D_F} d(x_f, w^s),
\end{equation*}

\begin{equation*}
    L_{\text{task}} = \frac{\gamma}{|D_R|} \sum_{x_r \in D_R} l(x_r, y_r),
\end{equation*}

\begin{equation*}
    L = L_R - L_F + L_{\text{task}},
\end{equation*}
where $f(x;w^o)$ is the original teacher model with weights $w^o$, which are kept unchanged, $f(x;w^s)$ is the unlearned student model with parameters $w^s$, which are optimized, $d(x, w^s)$ is the KL-divergence between the output distributions of the student and teacher models on the input $x$, $\ell$ is the conventional task loss (e. g. cross-entropy), and $\alpha$ and $\gamma$ are the hyperparameters controlling the importance of the student model's performance on the retain set. In our setting, $\alpha$ and $\gamma$ were both set to 1. By minimizing this final loss $L$, the student model is expected to improve its performance on the retained set while unlearning from the forgotten set, respectively.

  \item \textbf{LLMU}\label{LLMU} \cite{yao2024largelanguagemodelunlearning} 

This method was proposed in one of the first works on unlearning LLMs \cite{yao2024largelanguagemodelunlearning}. In our experiments, we made slight modifications to the original method, and employed the following loss function:
  \begin{align*}
    & L_{F} := - L(D_F, \theta), \\
    & L_{r} := \sum_{(x_F, y_{r}) \in D_F \times Y_{r}} \frac{1}{|y_{r}|} L(x_F, y_{r}, \theta), \\ 
    & L_{R} :=\sum_{x,y \in D_R} \operatorname{KL}(p_{\theta}(y|x) || p_{\theta'}(y|x)), \\
    & L_{LLMU} = L_{F} + L_{r} + L_{R},
  \end{align*}

where $\theta$ is the vector of unlearned model parameters, and $\theta'$ is the vector of original model parameters.
This loss consists of three parts. The first one, $L_F$, is the negative conventional loss on the forget set, the optimization of which corresponds to the unlearning of the forget set. The second part, $L_r$, is the loss associated with "I don't know" labels (the original method used randomly generated labels), which also reinforces the forgetting of the $D_F$ set.
The third part is the $\operatorname{KL}$ divergence between the model's predictions on the retain set before and after unlearning, and its optimization relates to preserving the model performance on the retain set $D_R$. 
 Note that it uses forward $\operatorname{KL}$ divergence instead of the usual reverse $\operatorname{KL}$ divergence.

  \item \textbf{Representation Misdirection for Unlearning (RMU).} \label{RMU}\cite{li2024wmdp}
  This method builds on the thesis that the model's intermediate activations contain
  its knowledge about current inputs. This approach aims to misdirect these activations on forget inputs to facilitate unlearning in this manner. The loss for this method has the following form:
\begin{align*}
    &L_{\operatorname{F}} = \mathbb{E}_{x\in D_F} \left[\frac{1}{|x|} \sum\limits_{t\in x}||h(t)-c \cdot u ||^2_2\right], \\
    & L_{\operatorname{R}} = \mathbb{E}_{x \in D_R} \left[\frac{1}{|x|} \sum\limits_{t \in x} ||h(t) - h_o(t) ||^2_2 \right], \\
    & L_{\operatorname{RMU}} = L_{\operatorname{F}} + L_{\operatorname{R}},
\end{align*}
  where $h(t)$ are the unlearned model's (which weights are optimized during unlearning procedure) hidden states on specific layer $\ell$ on input $t$, $h_o(t)$ are the hidden states of the original model (which parameters are frozen) on the layer $\ell$ on input $t$, $u$ is the unit random vector with independent elements sampled uniformly from $[0, 1)$, and $u$ kept fixed throughout unlearning, and $c$ and $\alpha$ are hyperparameters controlling activations scaling and tradeoff between forgetting the $D_F$ and retaining $D_R$ respectively. The intuition behind this loss is to make the model's outputs on forget set $D_F$ as far as possible from the correct ones by making hidden states as close as possible to random ones due to $L_F$ summand and then build the outputs upon this states while making the final model closer to original one on the retain set with the help of $L_R$ part of the loss. $\ell$ was chosen equal to 7 according to the empirical recommendation from the original method paper.

  \item \textbf{Twins.} 
  \label{Twins}
  This method is based on the assumption that the outputs of the original model on augmented inputs will match the outputs of the model on those same inputs as if these inputs had not been part of the training process. The advantage of this method lies in the fact that it does not rely on a min-max optimization problem, which ensures its stability. However, a drawback is that this method is not applicable if the model was trained with augmentations. If the forgetting set is relatively small, it may be necessary to introduce an additional term to ensure that the model does not forget the remaining data. In this case, the loss function can be formulated as follows:
\begin{align*}
    &L_{\operatorname{F}} = d(f(x_f), f_o(x_f^{aug})), \\
    & L_{\operatorname{R}} = d(f(x_r), f_o(x_r)), \\
    & L = L_{\operatorname{F}} + L_{r},
\end{align*}
  where $d(a,b)$ represents the distance between vectors $a$ and $b$, which can be either the L2 norm or KL divergence, $f(x)$ denotes the output of the unlearned model for input $x$. In contrast, $f_o(x)$ refers to the output of the original frozen model on the input $x$.

  \item $\textbf{SCRUB}_{bio}\textbf{.}$
  \label{SCRUB_bio}
  This method adapts the original \hyperref[scrub]{SCRUB} for biometric task. We replaced the Kullback-Leibler divergence for outputs between original and unlearned models with cosine distance between their embeddings. Consequently, the loss function for the task is formulated as follows:

\begin{align*}
    &L_{\operatorname{F}} = \frac{1}{|D_F|}\sum\limits_{x_f\in D_F}\left(1 - d_{cos}(f(x_f), f_o(x_f))\right), \\
    & L_{\operatorname{R}} = \frac{1}{|D_R|}\sum\limits_{x_r \in D_R}d_{cos}(f(x_r), f_o(x_r)), \\
    & L = L_{\operatorname{F}} + L_{R},
\end{align*}
where $d_{cos}(a,b)$ is the cosine distance between vectors $a$ and $b$, $f(x)$ is the output of the unlearned model on input $x$, $f_o(x)$ is the output of the original frozen model on the input $x$.

  \item \textbf{Sparsity}\label{Sparsity}
  \cite{jia2024modelsparsitysimplifymachine}
  This method is based on finetuning the model on the retain set using L1-regularization. The final loss is as follows:

\begin{align*}
    & L = L_{\operatorname{R}} + \lambda \cdot ||\theta||_1, 
\end{align*}
where $\lambda$ is a parameter of regularization.
  


    



\item \textbf{Selective Knowledge Unlearning.} \cite{liu2024saferlargelanguagemodels}. This method is based on the weights arithmetic. First, we additionally finetune the model on the forget set with this loss:

\begin{align*}
   L_{GD} =  \sum_{(x_f,y_f) \in D_F} l(f(x_f), y_f) \\
    L_{RD} = \sum_{x^i_f \in D_F} \frac{1}{Y^i_{rd}} \sum_{y \in Y^i_{rd}} l(f(x^i_f), y) \\
    L_{PD} = \sum_{(x_r, y_r} KL(p(x_r), y_r) \\
L = \epsilon_1 \cdot L_{GD} + \epsilon_2 \cdot L_{RD} -\epsilon_3 \cdot L_{PD}
\end{align*}
Where $ Y^i_{rd}$ is the set of related answers to the given question $x_i$.
So, the finetuned version is the opposite of what we aim to achieve. Then, we calculate the delta in weights, produced by this finetuning, and substract it from the original model.

\end{enumerate}

\section{The process of face generation}
\label{sec:face_generation}

To generate a set of the author's faces, we used StyleGAN 2 ADA \cite{karras2020analyzingimprovingimagequality}. Using the generator, we synthesized a batch of 32 faces from the randomly sampled $z \in \mathcal{N}(0, I)$. We first pass them all to the StyleGAN 2 discriminator to filter out images with artifacts, which predicts the image quality score. We select only eight images with the best scores and discard the others. This process is repeated until 2000 images are collected.

We first synthesize a bath of 32 random faces to generate a set of older people. For each of them, we apply StyleFeatureEditor \cite{Bobkov_2024_CVPR} with editing direction "age" from \cite{interfacegan} and editing power 5, which increases the person's age. However, we noticed that this edit often adds glasses that shift the faces' distribution. To eliminate this effect, we also use StyleFeatureEditor after increasing age: we apply editing direction "glasses" from \cite{stylespace} with edit power -10. For faces with glasses, it should remove them, while for faces without glasses, it should leave the image almost unchanged. Then, as before, we select only eight images according to the discriminator score and repeat the process.

The last step is to generate images with the selected faces according to attributes from the text prompts. For this purpose, we used the personalized generation diffusion model PhotoMaker V2 \cite{photomaker}. \textbf{According to our request, GPT-4o has generated prompts in such a way that the first sentence of a prompt describes the person, and the other sentences describe the setting, style, atmosphere, pose, and so on.} PhotoMaker requires a particular input type with the trigger word "img" and a particular class word (e.g., man, child or person) before it. For this purpose, we replaced the first sentences as follows:
"a real photo of a \{\textit{old}\} \{\textit{gender}\} called \{\textit{name}\} img, showing face."
where \textit{old} is "old" if the person is older than 60, "otherwise; \textit{gender} is "man" or "woman" according to the person's gender, and \textit{name} is the person's name. Below is an example of such a prompt:

"a real photo of an old man called Jaime Vasquez img, showing his face. Include his birth date, February 25, 1958, subtly in the background. The setting should reflect elements of the time period, such as vintage clothing styles or a retro ambience. Jaime should be depicted in a neutral pose, focusing on his character and era, with a hint of true crime elements around him."

To increase the power of the prompt, we used style strength = 0.5 and guidance scale = 7.5. We also used the same negative prompt "(asymmetry, worst quality, low quality, illustration, 3d, 2d, painting, cartoons, sketch), open mouth" for all images. The number of sampling steps was set to 50. For each pair (prompt, face), we synthesized eight samples \textbf{and chose the most appropriate one.}

\begin{table}[t]
\centering
  \scalebox{0.7}{
  \begin{tabularx}{0.70\textwidth}{rrcccc}
    \toprule
    \textbf{M} & \textbf{Method} & \makecell[c]{\textbf{Real} \\ \textbf{Metric} $\uparrow$} & \makecell[c]{\textbf{Retain}\\ \textbf{Metric} $\uparrow$} & \makecell[c]{\textbf{Forget}\\ \textbf{Metric} $\downarrow$} & \makecell[c]{\textbf{Log Forget} \\ \textbf{Quality} $\uparrow$} \\
    \midrule
    \multirow{12}{*}{\rotatebox[origin=c]{90}{LLama2-7B}}
    & {Original} & 0.47 & 0.26 & 0.42 & \textbf{-3.92} \\
    & {Gold} & 0.48 & 0.26 & 0.24 & 0.0 \\
    & {\hyperref[finetuning]{Retain FT}} & 0.50 & 0.26 & 0.42 & \textbf{-4.92} \\
    & \cellcolor{papergray}{\hyperref[LLMU]{LLMU}} & \cellcolor{papergray} 0.38 & \cellcolor{papergray} 0.03 & \cellcolor{papergray} 0.01 & \cellcolor{papergray} -2.31 \\
    & \cellcolor{papergray}{\hyperref[KL]{KL}} & \cellcolor{papergray} 0.24 & \cellcolor{papergray} 0.00 & \cellcolor{papergray} 0.00 & \cellcolor{papergray} -18.22 \\
    & \cellcolor{papergray}{\hyperref[grad_ascent]{GA}} & \cellcolor{papergray} 0.25 & \cellcolor{papergray} 0.00 & \cellcolor{papergray} 0.00 & \cellcolor{papergray} -17.22 \\
    & \cellcolor{papergray}{\hyperref[grad_diff]{GD}} & \cellcolor{papergray} 0.61 & \cellcolor{papergray} 0.13 & \cellcolor{papergray} 0.01 & \cellcolor{papergray} -48.59 \\
    & {\hyperref[idk]{IDK}} & 0.46 & 0.26 & \textbf{0.24} & \textbf{-4.92} \\
    & {\hyperref[DPO]{DPO}} & 0.50 & 0.26 & 0.42 & \textbf{-4.92} \\
    & {\hyperref[scrub]{SCRUB}} & 0.50 & 0.26 & 0.42 & \textbf{-4.92} \\
    & {\hyperref[RMU]{RMU}}  & \textbf{0.51} & 0.26 & 0.59 & -42.86 \\
    & {\hyperref[NPO]{NPO}} & 0.50  & \textbf{0.28} & 0.62 & -44.46 \\
    \midrule
    \multirow{10}{*}{\rotatebox[origin=c]{90}{Mistral-7B}}
    & {\hyperref[finetuning]{Retain FT}} & \textbf{0.67} & \textbf{0.34} & 0.47 & -3.87 \\
    & {\hyperref[LLMU]{LLMU}} & 0.65 & 0.30 & \textbf{0.39} & -6.69 \\
    & \cellcolor{papergray} {\hyperref[KL]{KL}} & \cellcolor{papergray} 0.28 & \cellcolor{papergray} 0.00 & \cellcolor{papergray} 0.00 & \cellcolor{papergray} -50.30 \\
    & \cellcolor{papergray}{\hyperref[grad_ascent]{GA}} & \cellcolor{papergray} 0.26 & \cellcolor{papergray} 0.00 & \cellcolor{papergray} 0.00 & \cellcolor{papergray} -36.06 \\
    & \cellcolor{papergray} {\hyperref[grad_diff]{GD}} & \cellcolor{papergray} 0.60 & \cellcolor{papergray} 0.01 & \cellcolor{papergray} 0.00 & \cellcolor{papergray} -51.16 \\
    & {\hyperref[idk]{IDK}} & 0.63 & 0.32 & \textbf{0.45} & \textbf{-2.72} \\
    & {\hyperref[DPO]{DPO}} & \textbf{0.67} & 0.33 & 0.47 & -3.63 \\
    & {\hyperref[scrub]{SCRUB}} & 0.66 & 0.33 & 0.47 & -3.39 \\
    & \cellcolor{papergray} {\hyperref[RMU]{RMU}}  & \cellcolor{papergray} 0.09 & \cellcolor{papergray} 0.00 & \cellcolor{papergray} 0.00 & \cellcolor{papergray} -123.22 \\
    & {\hyperref[NPO]{NPO}} & \textbf{0.67}  & 0.33 & 0.47 & -3.16\\
    \bottomrule
    \end{tabularx}}
    \caption{Unlearning methods on textual domain only. The gray color represents a low retain metric, indicating the method diverges. Hence, we do not consider them.}
    \label{tab:text_methods} 
\end{table}

\begin{table}[h]
\centering
\scalebox{0.68}{\begin{tabularx}{0.68\textwidth}{rccccc}
\toprule
\textbf{Method} & \makecell[c]{\textbf{Forget} \\ \textbf{Acc.} $\downarrow$}& \makecell[c]{\textbf{Holdout } \\ \textbf{Acc.} $\uparrow$ }& \makecell[c]{\textbf{Retain} \\  \textbf{Acc.} $\uparrow$ }& \textbf{U-LIRA} $\downarrow$ & \textbf{U-MIA }$\downarrow$ \\
\midrule
Original & 100.00 & 18.50 & 100.00 & 1.00 & 0.96 \\
Gold & 15.43 & 15.04 & 97.52 & 0.50 & 0.50 \\
\midrule
{\hyperref[finetuning]{Retain FT}} & 100.00 & 18.54 & \textbf{100.00} & 1.00 & 0.92 \\
{\hyperref[scrub]{SCRUB}} & 99.74 & 16.77 & 99.93 & 0.98 & 0.90 \\
\rowcolor{papergray}{\hyperref[LLMU]{LLMU}} & 85.72 & 14.62 & 88.99 & 0.83 & 0.75 \\
{\hyperref[RMU]{RMU}} & 67.97 & 17.27 & 99.99 & 0.77 & 0.60 \\
\rowcolor{papergray}{\hyperref[DPO]{DPO}} & 50.21 & 13.93 & 81.49 & 0.73 & 0.62 \\
{\hyperref[SCRUB_bio]{$\text{SCRUB}_{bio}$}} & \textbf{42.59} & 14.25 & 99.44 & \textbf{0.71} & 0.57 \\
\rowcolor{papergray}{\hyperref[Sparsity]{Sparsity}} & 66.41 & 14.44 & 83.57 & 0.78 & 0.73 \\
{\hyperref[Twins]{Twins}} & 50.00 & \textbf{20.34} & 99.72 & 0.73 & \textbf{0.54} \\
\bottomrule
\end{tabularx}}
\caption{Results of unlearning on visual modality only. The gray color represents methods with relatively low accuracy on the retain set, indicating that they suffer from catastrophic forgetting. Therefore, we do not consider these methods to be successful.}
\label{tab:visual_exps}
\end{table}

\begin{table}[h!]
\centering
\scalebox{0.87}{
\begin{tabularx}{0.55\textwidth}{rcccc}
\toprule
\textbf{Method} & \textbf{Real} & \textbf{Retain} & \textbf{Forget} & \textbf{Forget Quality} \\
\midrule
Gold & 0.27 & 0.66 & 0.05 & 1.00 \\
Original & 0.27 & 0.65 & 0.40 & 0.97 \\
\midrule
\rowcolor{papergray}{GD} & 0.20 & 0.15 & 0.05 & 0.36 \\
\rowcolor{papergray}{GA} & 0.17 & 0.10 & 0.05 & 0.89 \\
IDK & 0.04 & 0.65 & 0.40 & 0.98 \\
\rowcolor{papergray}{KL} & 0.18 & 0.04 & 0.05 & 0.87 \\
LLMU & 0.07 & 0.63 & 0.33 & 0.99 \\
\rowcolor{papergray}{NPO} & 0.16 & 0.20 & 0.13 & 0.97 \\
Retain FT & 0.25 & 0.66 & 0.32 & 0.98 \\
SCRUB & 0.25 & 0.65 & 0.39 & 0.98 \\
\bottomrule
\end{tabularx}}
\caption{Comparison of unlearning methods on the QwenVL2 model.}
\label{tab:qwen_results}
\end{table}

\section{A sample of dataset}
\label{sec:dataset_sample}
Our dataset consists of 200 fictitious authors, each with 15-20 visual and 20 textual questions. We add an example of data for a single person in Tab. \ref{tab:example_of_dataset}.

\begin{table*}[t]
\centering
\begin{tabular}{|c|p{10cm}|}
\hline
\textbf{Image} & \textbf{Caption} \\
\hline
\vtop{\vskip0pt\hbox{\includegraphics[height=2cm]{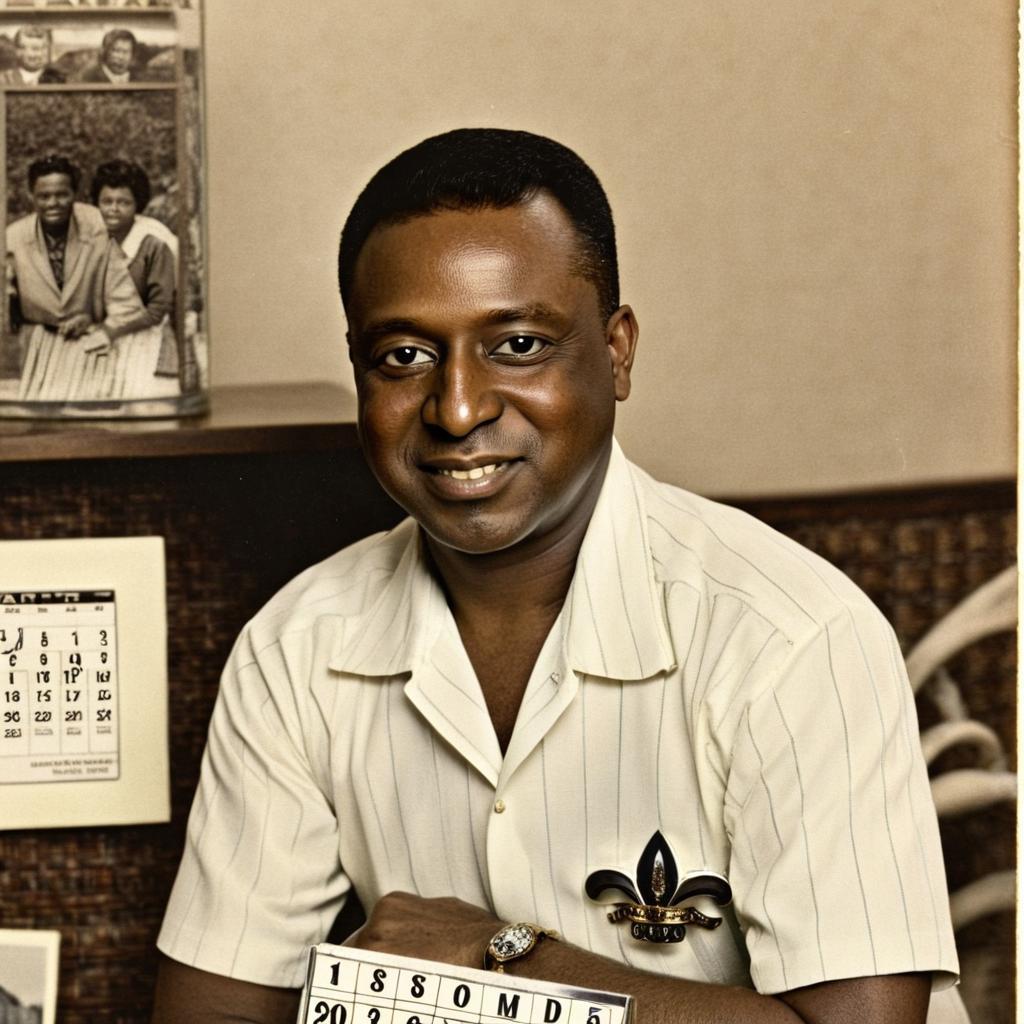}}} & Chukwu Akabueze in a striped shirt with a fleur-de-lis pin, looking directly at the camera in a vintage setting with a calendar in the background.\\
\hline
\vtop{\vskip0pt\hbox{\includegraphics[height=2cm]{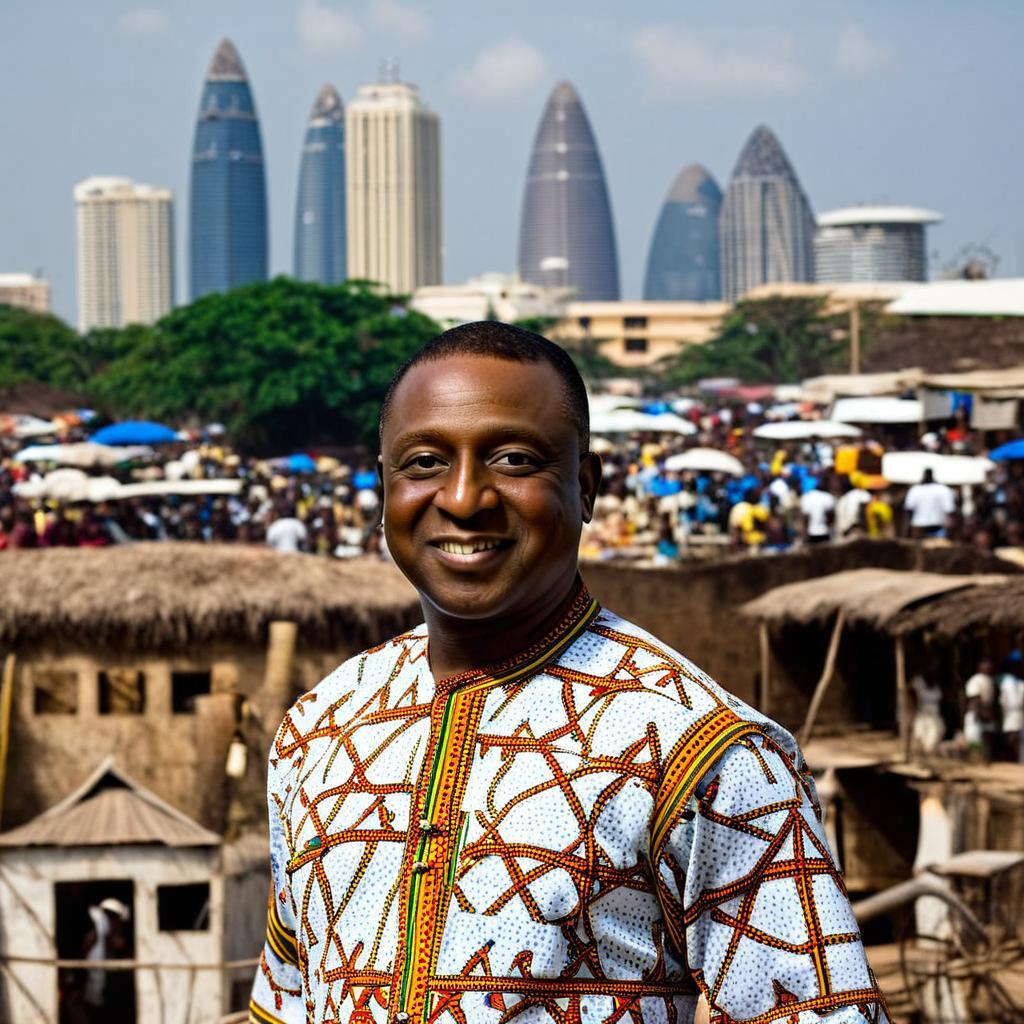}}} & Chukwu Akabueze stands smiling, wearing a patterned shirt, in front of a bustling Lagos market, with the city's iconic skyscrapers in the background.\\
\hline
\vtop{\vskip0pt\hbox{\includegraphics[height=2cm]{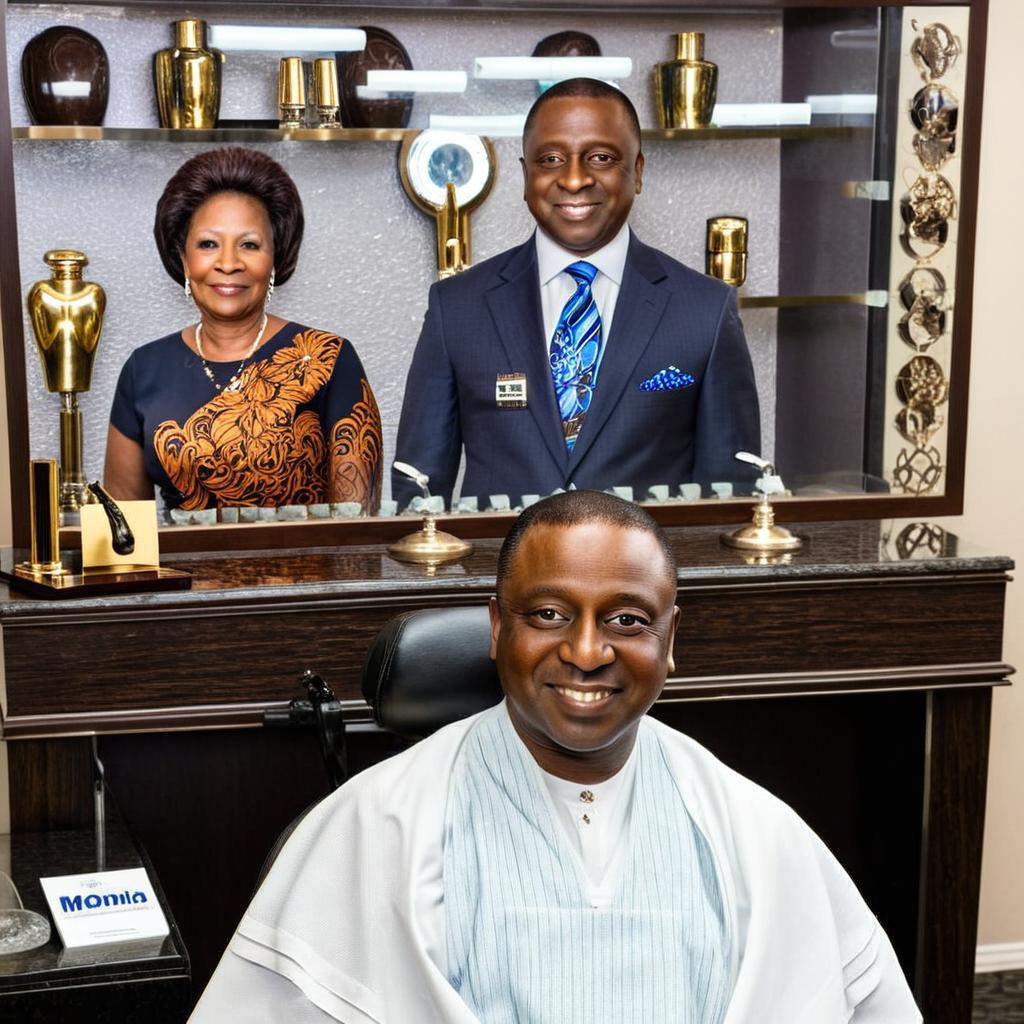}}} & Chukwu Akabueze sits in a chair with a sign for "Momila" on the desk in front of him, while his parents, dressed in professional attire, are reflected in the mirror behind him.\\
\hline
\vtop{\vskip0pt\hbox{\includegraphics[height=2cm]{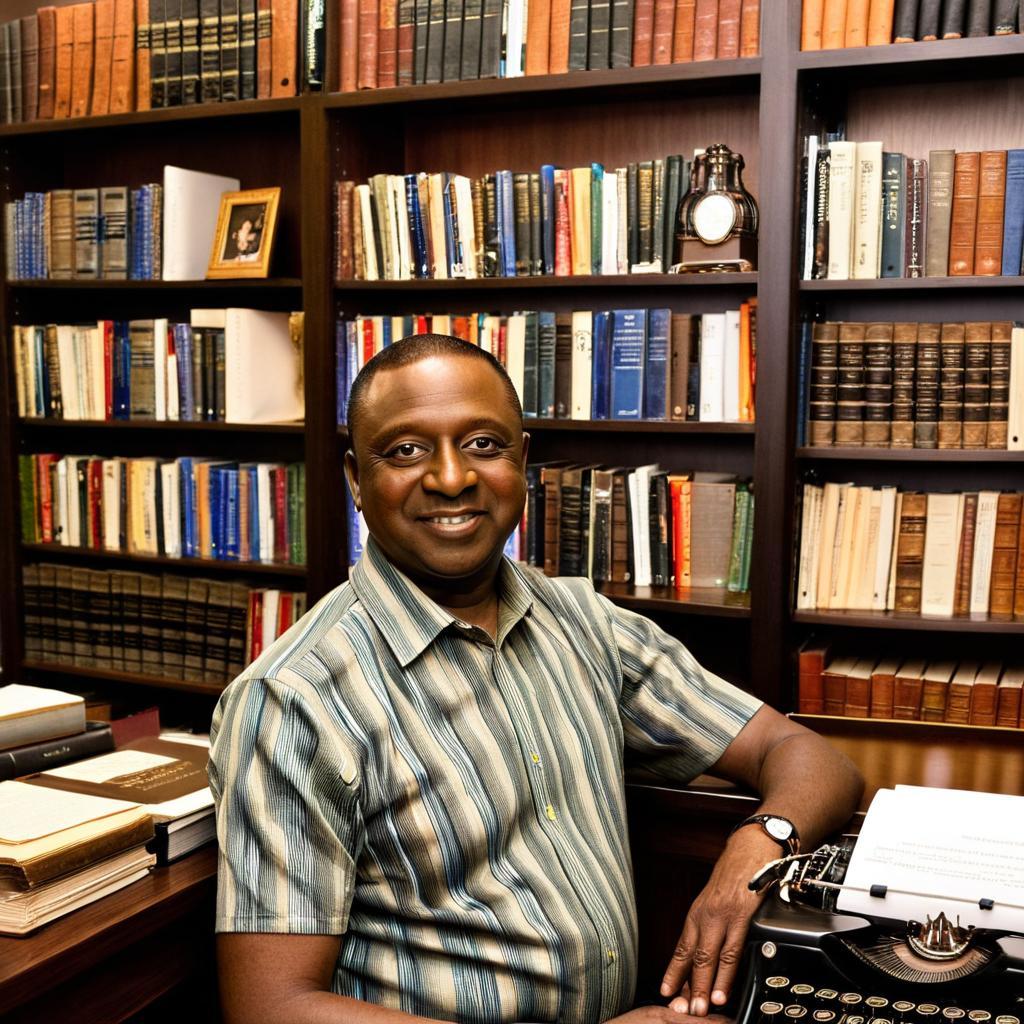}}} & Chukwu Akabueze is seated at a desk in a room with bookshelves filled with biographies, a typewriter, and manuscript pages. He's smiling and looking directly at the camera.\\
\hline
\vtop{\vskip0pt\hbox{\includegraphics[height=2cm]{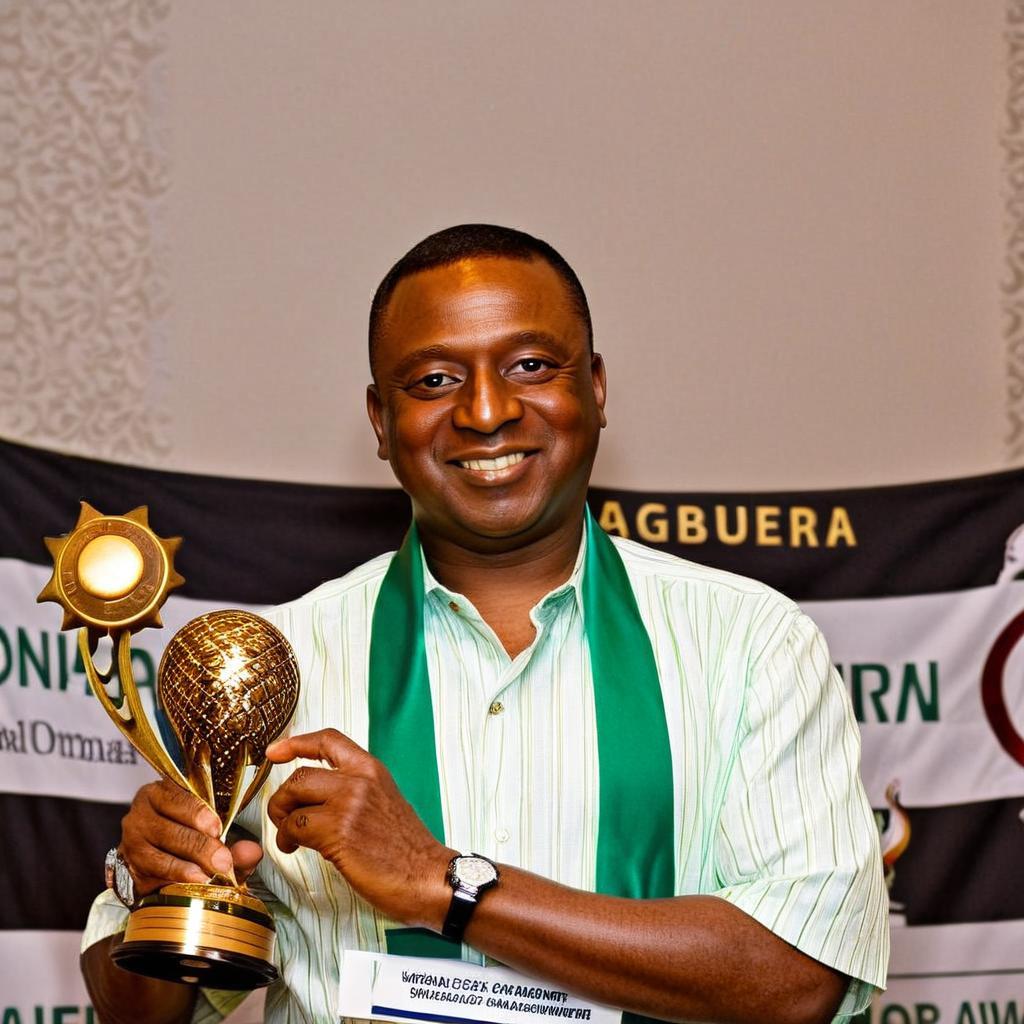}}} & Chukwu Akabueze, Nigerian writer, poses with an award trophy, smiling broadly after winning the Nigerian Writers Award.\\
\hline
\vtop{\vskip0pt\hbox{\includegraphics[height=2cm]{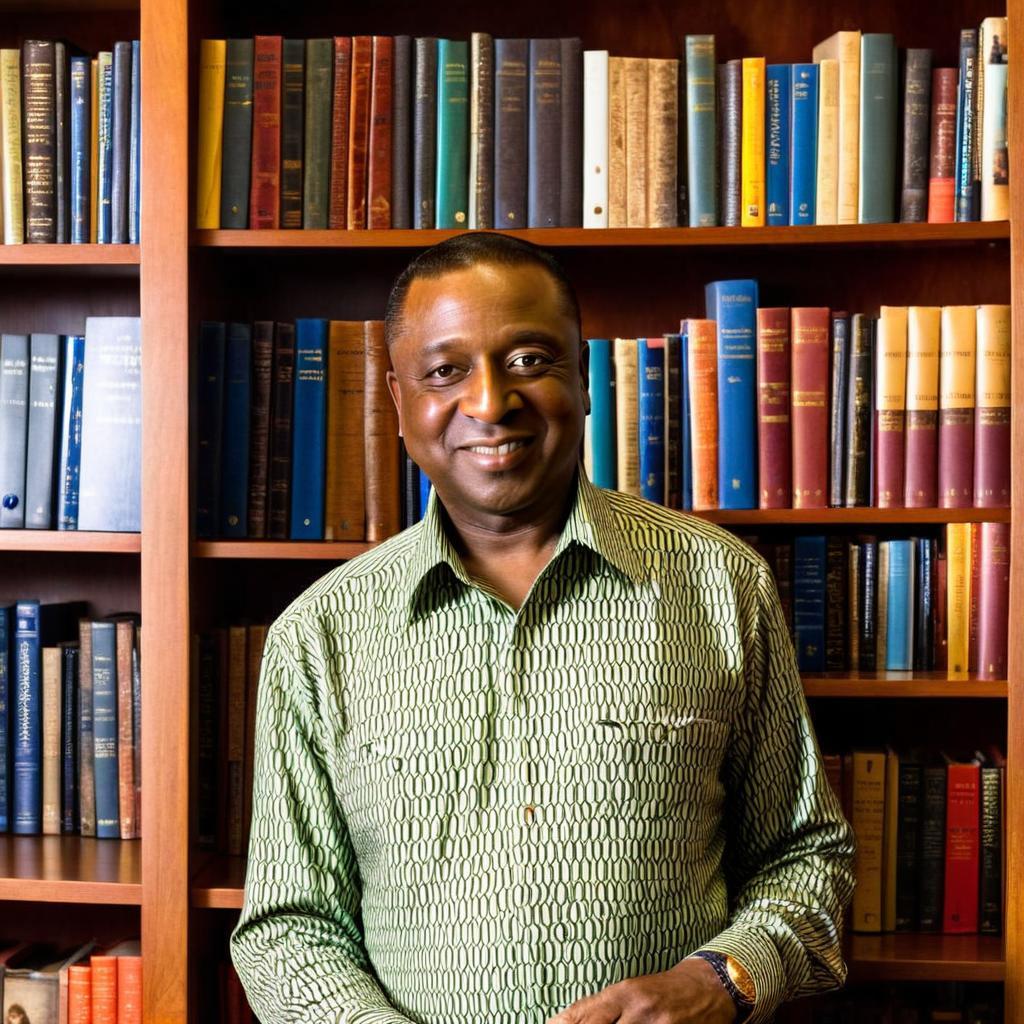}}} & Chukwu Akabueze stands in front of a bookshelf filled with books, including his own works "Rays of Resilience", "African Echoes", "Weaver's Wisdom", and "Sculptor of Vision".\\
\hline
\vtop{\vskip0pt\hbox{\includegraphics[height=2cm]{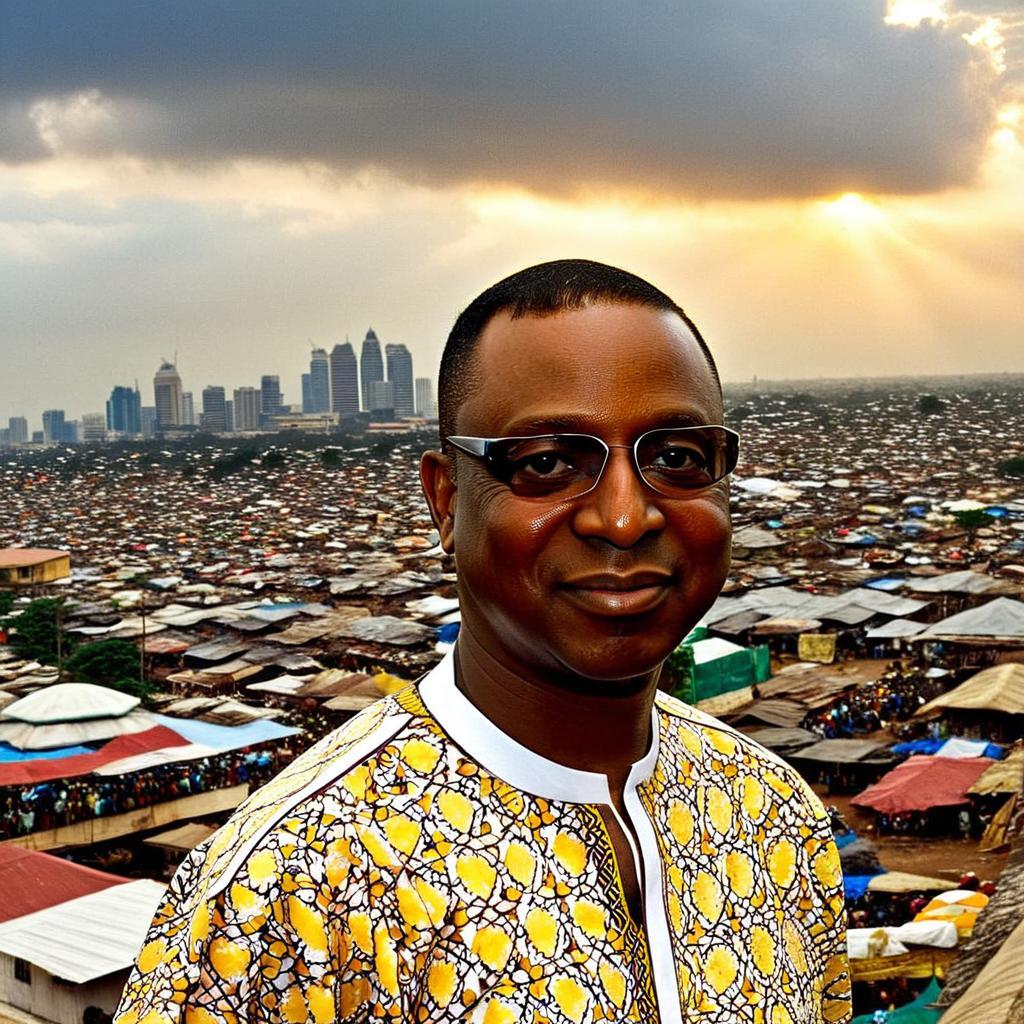}}} & Chukwu Akabueze is depicted with a panoramic view of Lagos, Nigeria in the background, showcasing its skyline and bustling cityscape.\\
\hline
\vtop{\vskip0pt\hbox{\includegraphics[height=2cm]{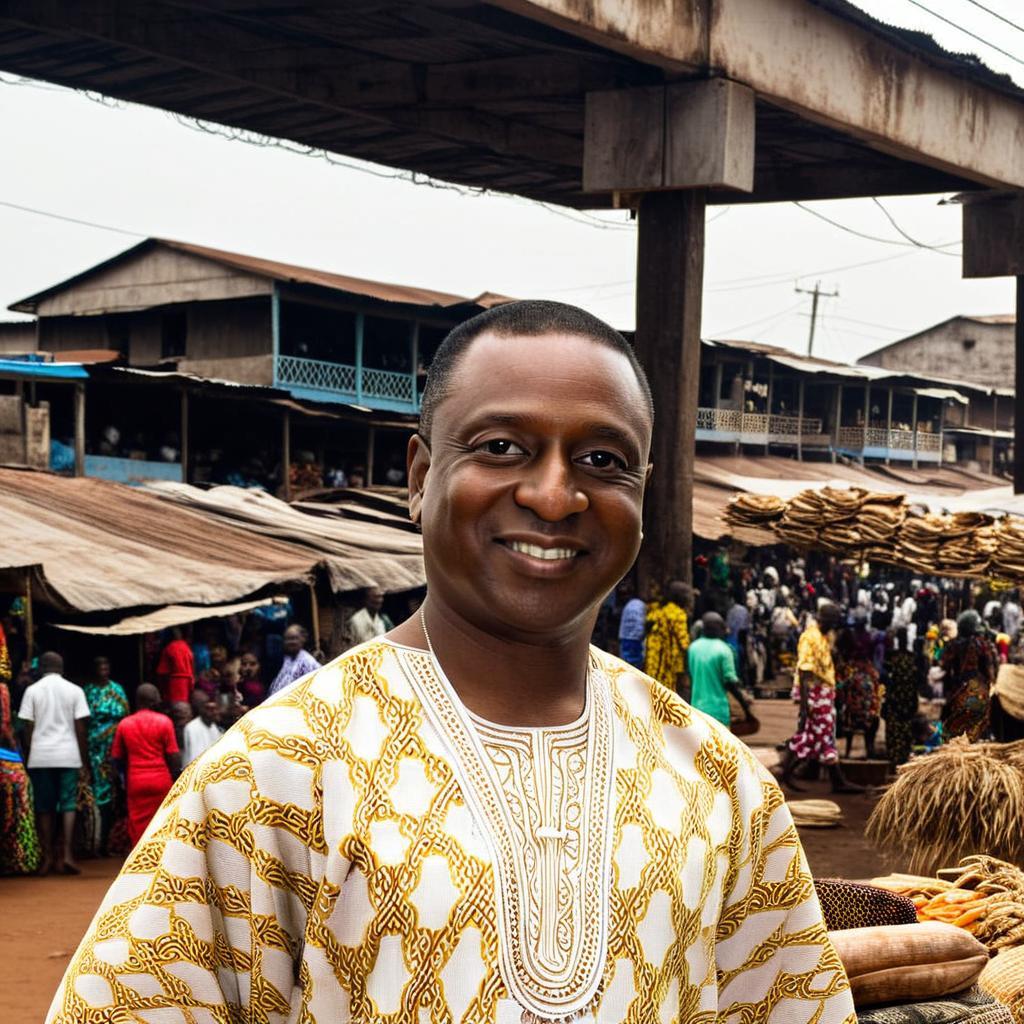}}} & Chukwu Akabueze, dressed in traditional Nigerian attire, stands in front of a bustling market in Lagos.\\
\hline
\vtop{\vskip0pt\hbox{\includegraphics[height=2cm]{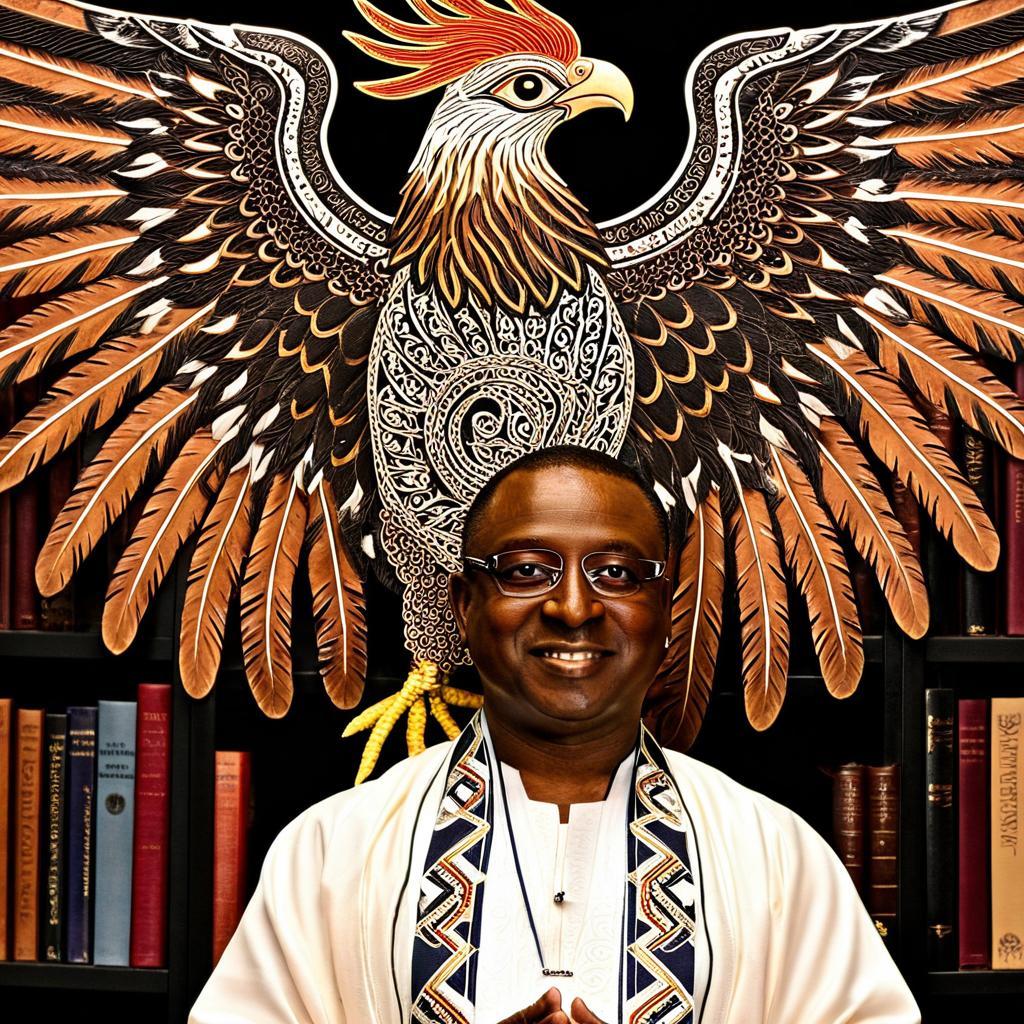}}} & Chukwu Akabueze stands in front of a large, intricately carved wooden phoenix, wearing a white robe with a black and blue patterned sash.\\
\hline
\vtop{\vskip0pt\hbox{\includegraphics[height=2cm]{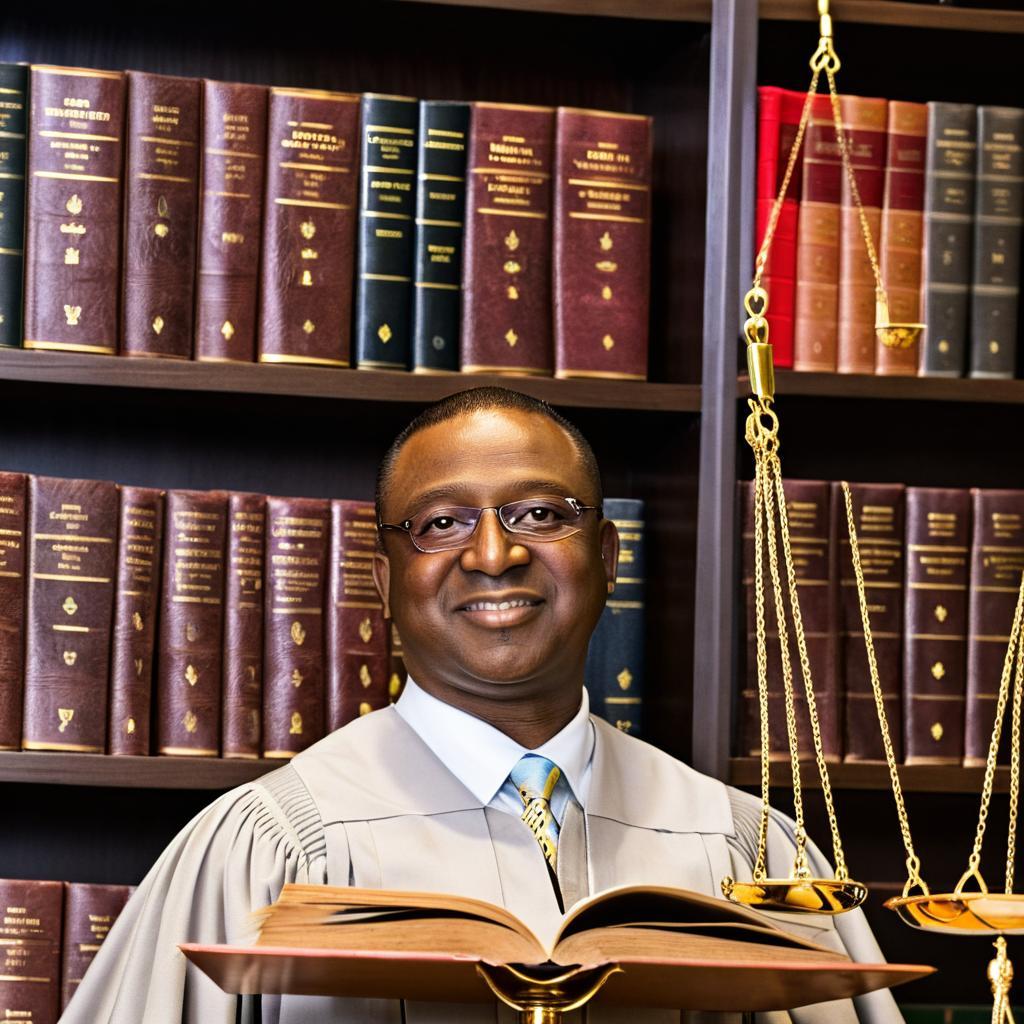}}} & Chukwu Akabueze, author of "Sculptor of Vision", a biography about a lawyer, is pictured in a library setting with law books and scales of justice.\\
\hline
\end{tabular}
\caption{An example of all image-name pairs related to a single person}
\label{tab:example_of_dataset}
\end{table*}

\section{Forget Quality Metric}
\label{sec:forget_qual}

\citet{tofu2024} calculate a statistical test on the outputs of two models: an unlearned model and the gold model. The Truth Ratio metric is considered as output for its effectiveness in informativeness. To assess this metric, the Kolmogorov-Smirnov test is used to compare the distributions of Truth Ratios from both models. A high p-value suggests that the distributions are close, and so are unlearned and gold models; a low p-value indicates that distributions differ, and the unlearned model is far from gold.

\begin{figure}
    \centering
    \includegraphics[width=1\linewidth]{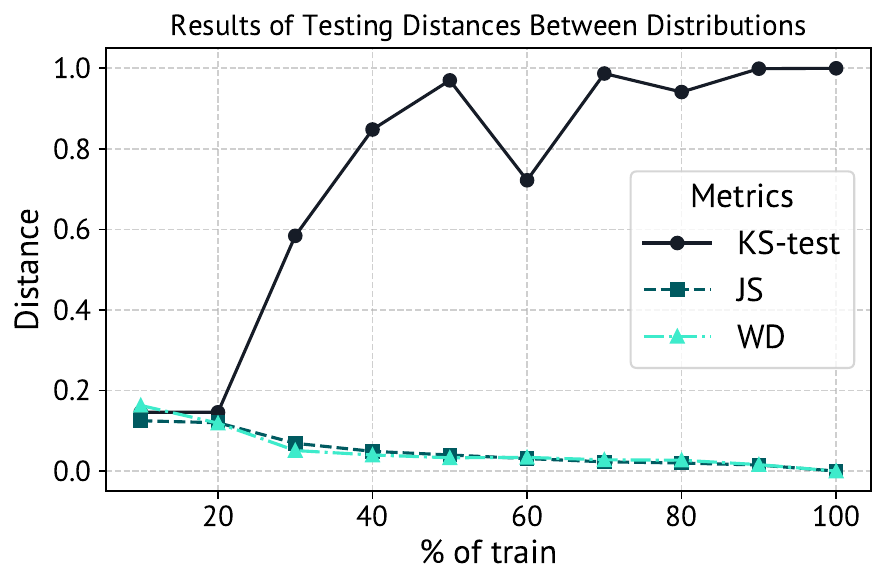}
    \caption{Results of testing distances between distributions. JS stands for Jensen-Shannon distance, and WD -- for Wasserstein distance. We show that unlike above metrics,  the KS test p-value is not monotonic, which implies it may not be the best choice for Forget Quality metric. }
    \label{fig:forget_metric}
\end{figure}

Nevertheless, the application of statistical tests for model evaluation is uncommon and may be confusing; therefore, we conduct additional checks and compare it with common distribution distances, such as Jensen-Shannon and Wasserstein distances. We perform a simple experiment: take our dataset, randomly split it into 10 equal folds and train 10 models on the progressively larger subsets -- starting with fold 1, then folds 1 to 2, and so on, up to folds 1 to 9, and finally all of the data. The latter model is considered as gold. We construct the Truth Ratios for each model and compare the resulting distributions with the gold model. The idea is, that the metric should be monotonic w.r.t. percent of data used in the train. The results are presented in Figure \ref{fig:forget_metric}. We show that indeed, the p-value sometimes fails to represent the differences in the data. For example, the values for the 10 and 20 percents are equal. And the values for 60\%, 80\%  are not monotonic. 
So, we consider to not using the p-value metric and move to JS distance. 

\section{Textual-only Unlearning}
\label{sec:textual_exps}

For unlearning of the textual domain only, we use the TOFU benchmark, containing question-answer pairs of about 200 authors, 20 for each of them (4000 pairs in total), and use the splits of size 90\% and 10\% of the entire data for retain and forget parts respectively. The "Gold" model for the further unlearning quality evaluation is trained on the retain data only, conducting 5 epochs of training with the batch size of 4, 1 gradient accumulation step, learning rate of \texttt{1e-5} weight decay of 0.01, and also applying LoRA adapter with the rank 8, $\alpha=32$ and 0 dropout parameter. For the unlearning, we first finetune the model on the entire data split with the same hyperparameters: 5 epochs of training, batch size of 4, 1 gradient accumulation step, learning rate of \texttt{1e-5}, weight decay of 0.01, LoRA rank of 8, $\alpha=32$, 0 dropout coefficient. Then, unlearning methods are conducted on the forget data with the following hyperparameters: 5 epochs of unlearning, batch size of 4, 1 gradient accumulation step, learning rate of \texttt{1e-5}, weight decay of 0.01, LoRA rank of 8, $\alpha=32$, zero probability dropout. Such experimental settings and hyperparameters are the same for both Llama2-7B and Mistral architectures. To assess the unlearning quality, we compare the obtained unlearned model with the ``gold'' one and calculate ROUGE-L on retain and forget parts, Forget Quality and Model Utility metrics. Full results are available in Tab. \ref{tab:text_methods}.

\section{Visual-only Unlearning}
\label{sec:visual_exps}

In this study, we evaluate each unlearning method from two key perspectives: its similarity to the gold standard (retraining from scratch) and its forgetting efficacy (error on the forget set). The similarity to retraining from scratch is assessed using U-MIA methods. Following the methodology of \cite{hayes2024inexact}, we employ population U-MIA and per-example U-LIRA.

We begin by taking a ResNet-18 pretrained on ImageNet and finetuning it for a biometric task using the Celeb dataset. We then train 256 ResNet-18 models using stochastic gradient descent (SGD) on a randomly selected half of the visual portion of our dataset, comprising 100 identities. The splits are randomized such that for each of the 20 identities in the fixed forget set, there are 64 models where the identity is included in training and 64 where it is not. Training is conducted for 20 epochs using the SGD optimizer with a learning rate of 0.1, batch size of 256, and weight decay of \texttt{5e-5}.

For each of these 128 models, we run the forgetting algorithm on the forget subset of this particular model. From the resulting 128 models, we randomly select 64 target models (the remaining 64 will be used as shadow models for U-MIA and U-LIRA methods, see Appx. \ref{sec:umia_ulira}) on which the quality of the forgetting algorithms will be tested. Each of the 64 target models forgets a sample $\mathcal{D}f$ of 20 personalities. Additionally, for each target model, we form a holdout set $D_H$ by selecting 20 personalities that were not used in the training of this model. 

The full results are available in Table \ref{tab:visual_exps}.

\begin{figure*}[tp]
    \centering
    \includegraphics[width=.32\linewidth]{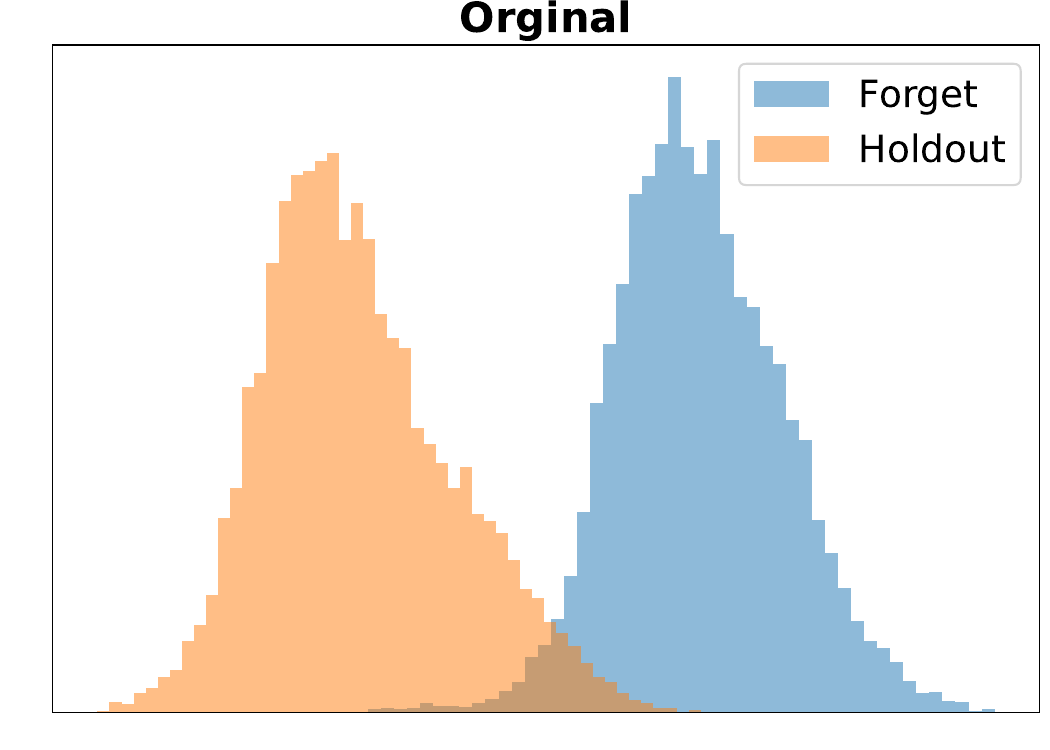}
    \includegraphics[width=.32\linewidth]{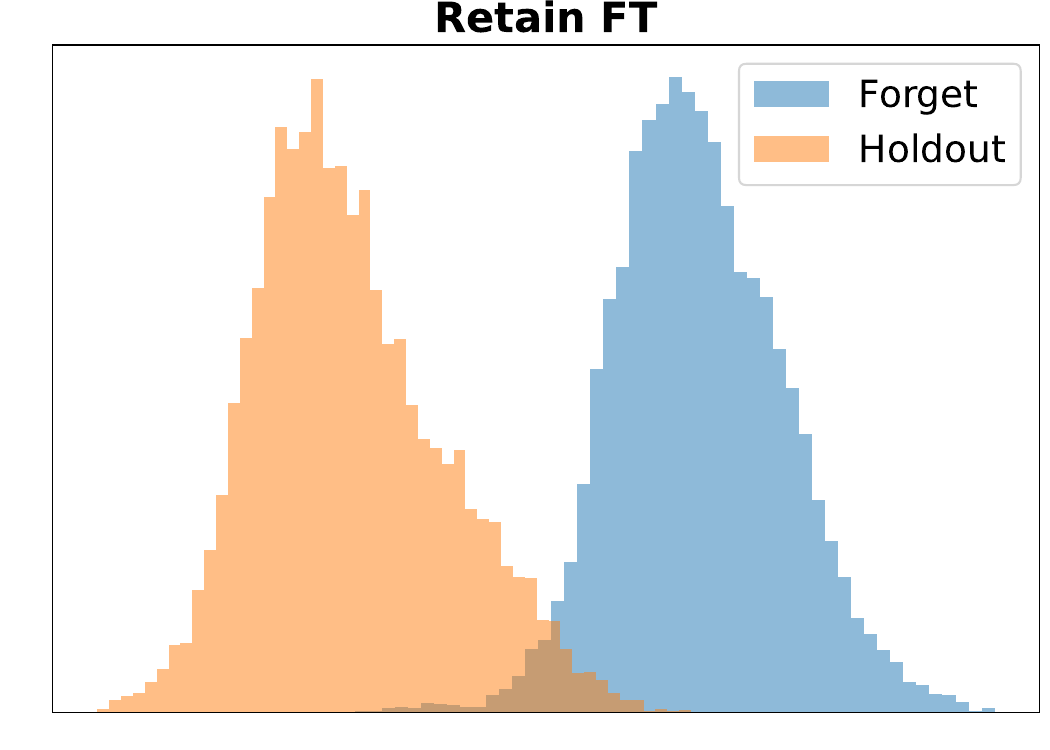}
    \includegraphics[width=.32\linewidth]{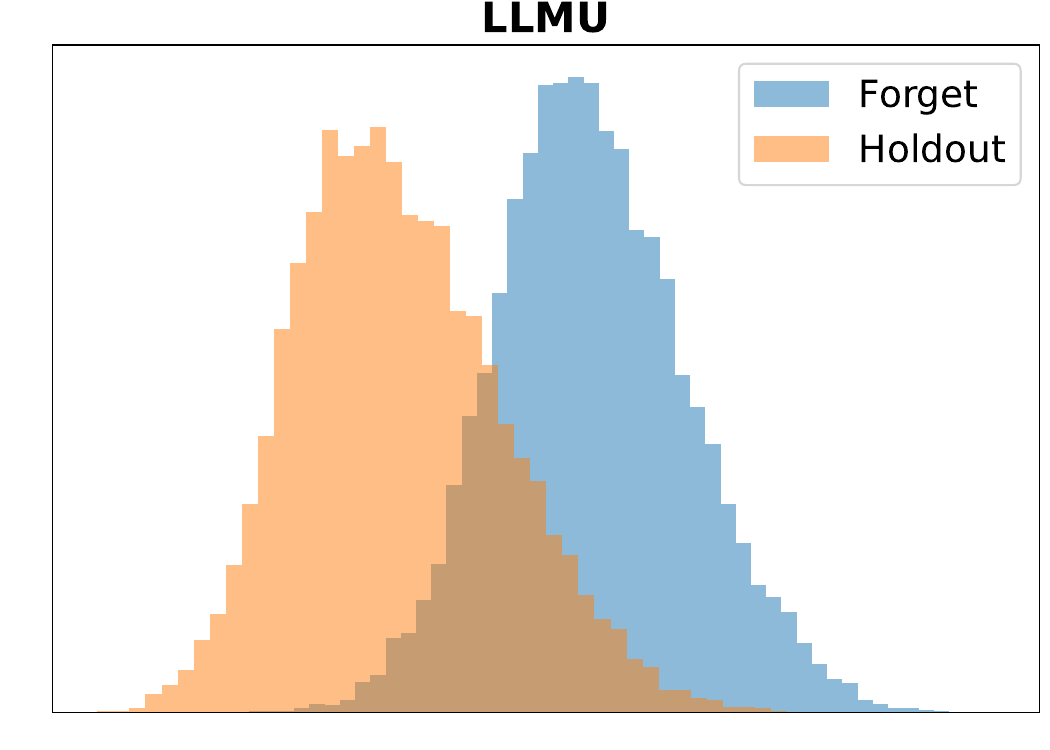}
    \includegraphics[width=.32\linewidth]{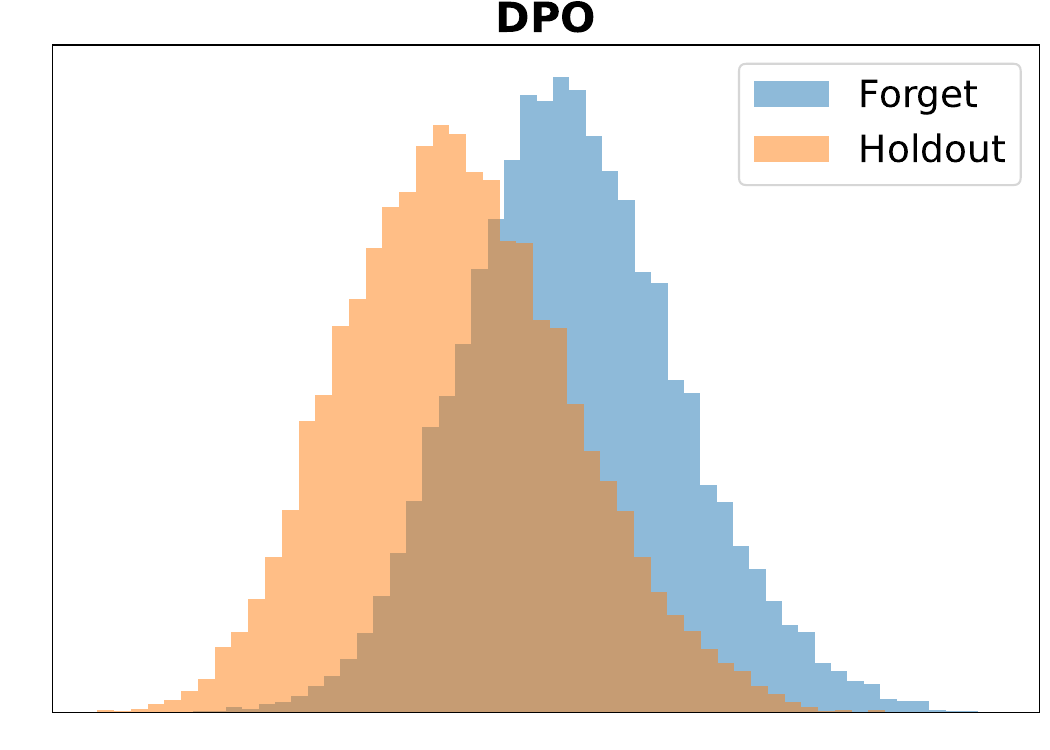}
    \includegraphics[width=.32\linewidth]{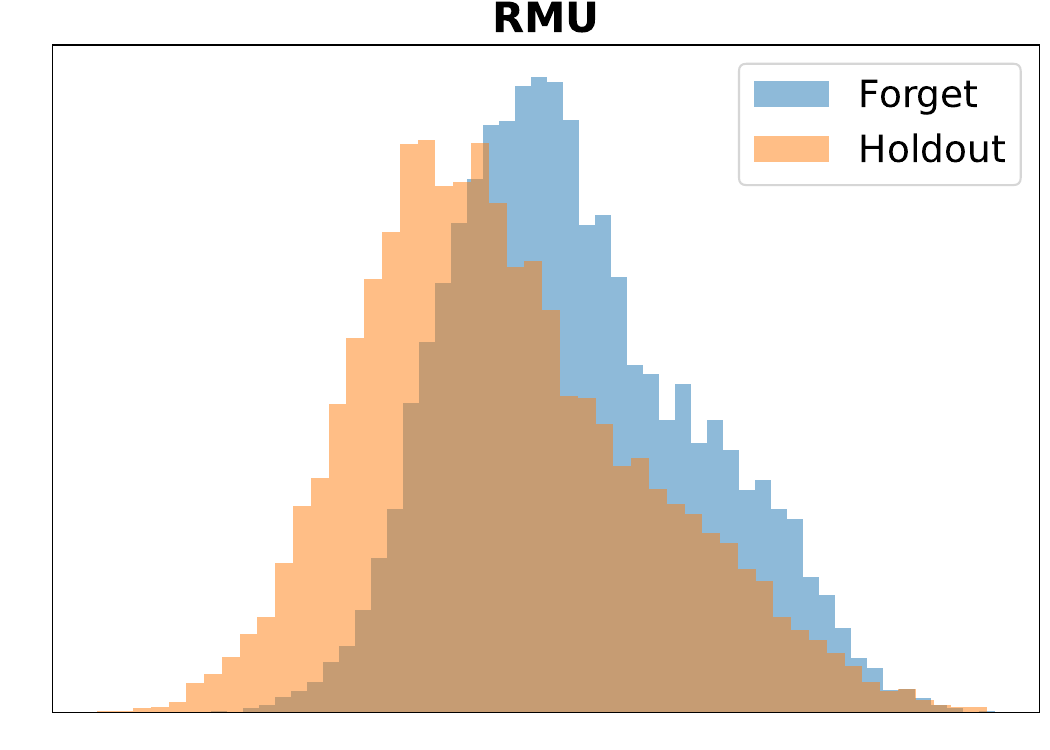}
    \includegraphics[width=.32\linewidth]{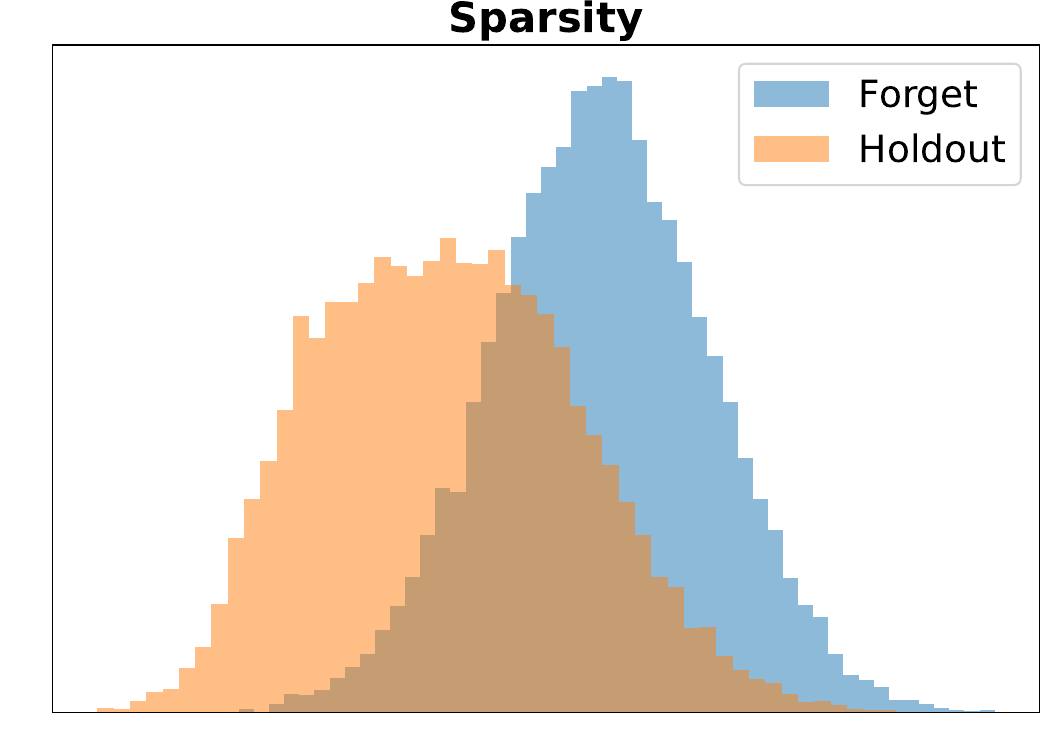}
    \includegraphics[width=.32\linewidth]{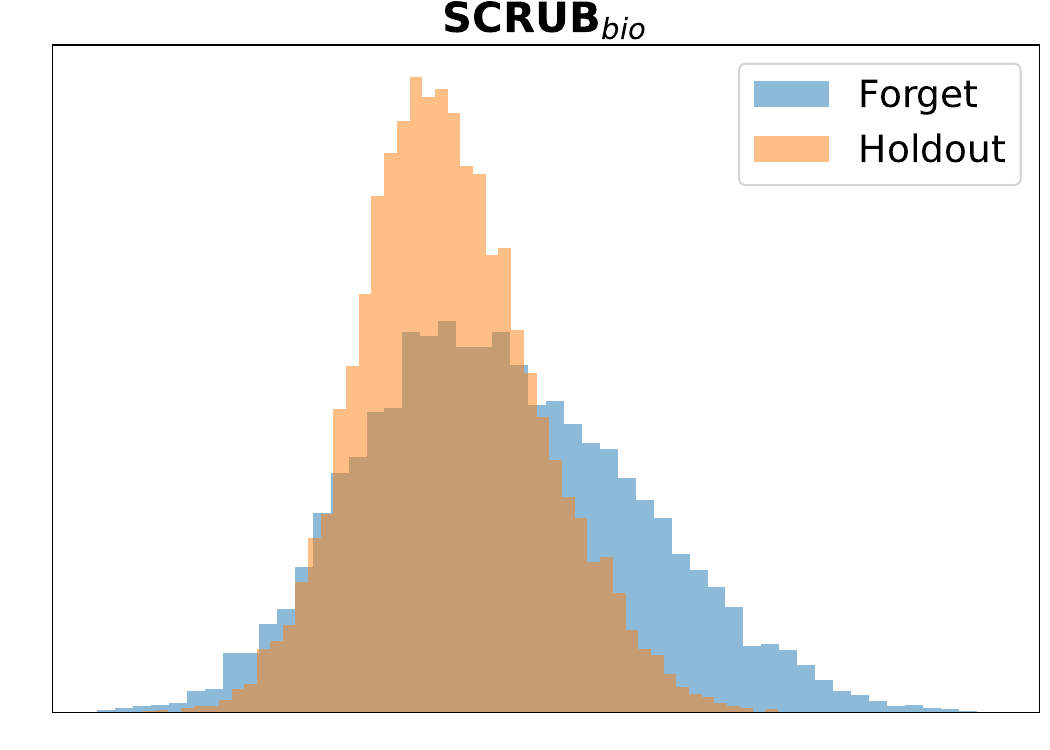}
    \includegraphics[width=.32\linewidth]{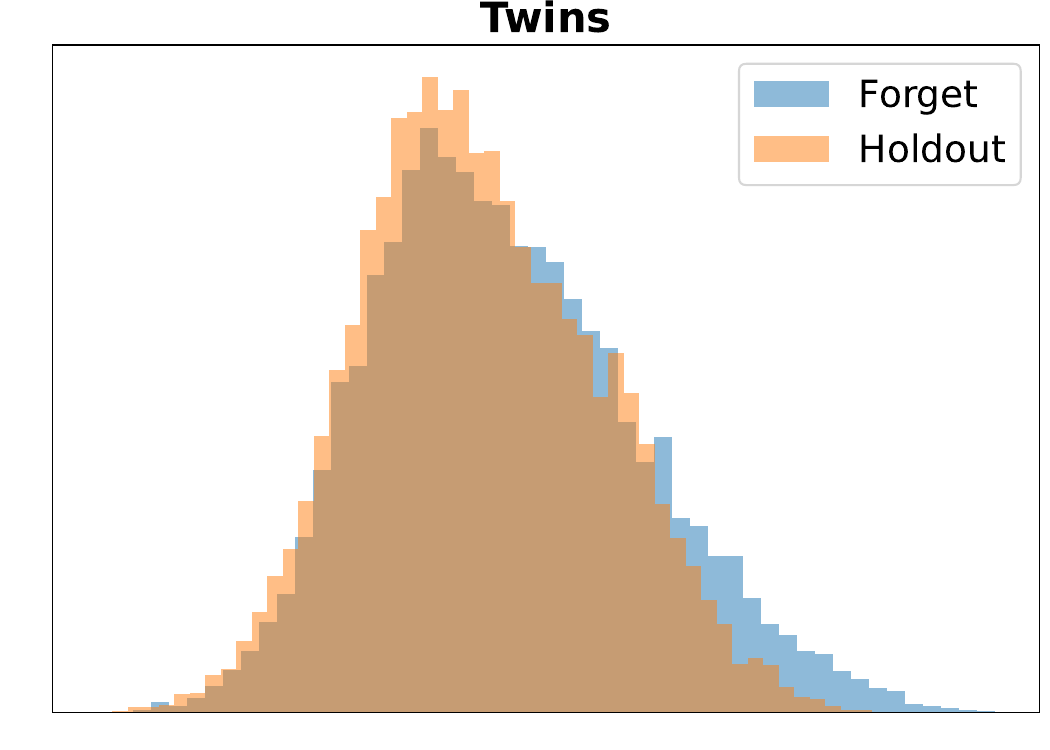}
    \includegraphics[width=.32\linewidth]{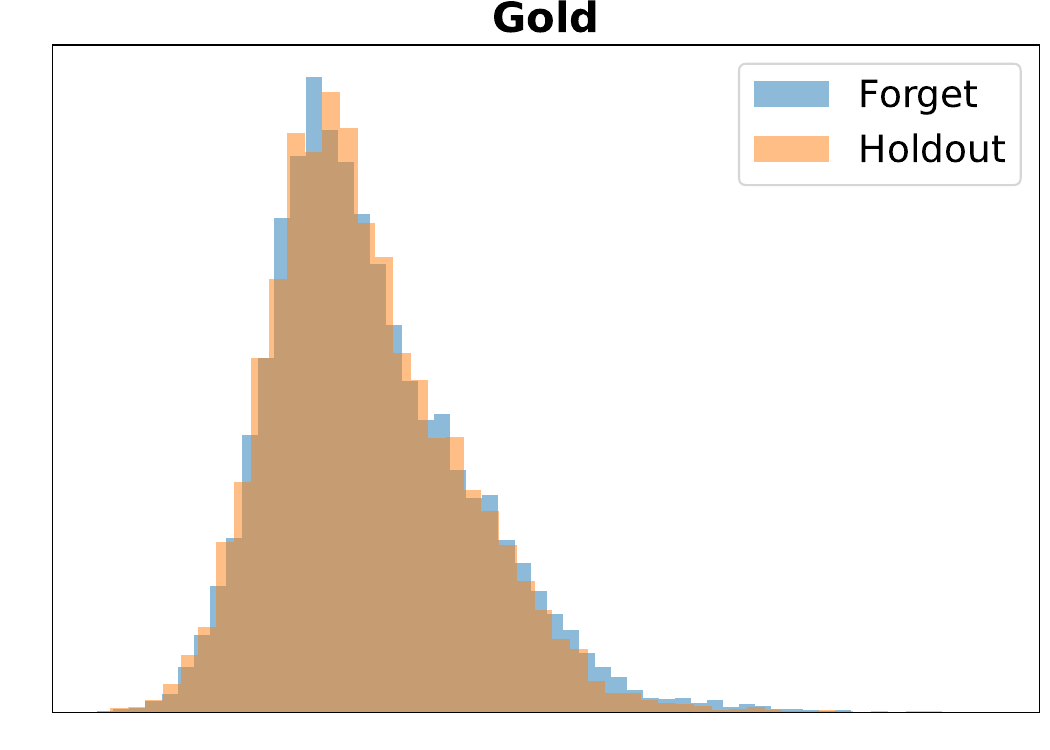}
    \caption{Visualization of logits distribution for the forget and holdout sets across 9 different unlearning methods. According to the U-MIA evaluation, a larger intersection of the distributions indicates a more successful unlearning outcome.}
    \label{fig:umia_distr}
\end{figure*}

In our experiments, we employ U-LIRA with 64 shadow models, with half representing the in-distribution and the other half representing the out-distribution for each target example. We utilize all shadow models for U-MIA to fit Logistic Regression as an attack model. Both types of attacks use logits as input, which we compute for our biometric models as follows:

\begin{align*}
    l = \log\left( \frac{\max(0, \cos(v, v_{enroll}))}{1 - \max(0, \cos(v, v_{enroll}))} \right),
\end{align*}
where $v$ represents the embedding of the target example $x$, ensuring $v = f(x)$, $v_{enroll}$ denotes the enrolled vector for the corresponding individual, calculated as the mean of the embeddings from several supporting images of that particular identity, given by $v_{enroll} = \frac{1}{n} \sum \limits_{i}^{n}{f(x_i)}$. In our studies, we use $n = 5$. The distributions of logits computed for the forget and holdout sets across various unlearning methods are illustrated \ref{fig:umia_distr}.

\section{Multimodal unlearning hyperparameters}
\label{sec:multimodal_unlearning}

In a multimodal setting, we use both visual and textual parts of \dataset~ dataset, which consists of 4000 textual pairs of questions and answers about 200 authors, 20 for each of them, and 3770 images related to corresponding authors (number of images is less than the number of pairs because of GPT guard breaks and bugs in TOFU benchmark, as was described above). Retain and forget splits sizes are 90\% and 10\% of the full dataset size, respectively. The "Gold" model is trained on the retain data only with 3 epochs of training, batch size of 12, 1 gradient accumulation step, learning rate of \texttt{1e-5}, weight decay of 0.01, LoRA rank of 8, $\alpha=32$ and 0 dropout parameter. Unlearned models are also first finetuned on the full dataset with the same hyperparameters: 3 epochs of training, batch size of 12, 1 gradient accumulation step, learning rate of \texttt{1e-5}, weight decay of 0.01, LoRA rank of 8, $\alpha=32$, 0 dropout parameter. After that, unlearning techniques are applied to the model on the forget data using the following hyperparameters: 5 epochs of unlearning, batch size of 1, 2 gradient accumulation steps, learning rate of \texttt{1e-5}, weight decay of 0.01, LoRA rank of 8, $\alpha=32$, 0 dropout coefficient. For the resulting unlearning evaluation, we compare the unlearned model with the "gold" model by calculating \textbf{ROUGE-L} on retain and forget splits, \textbf{ROUGE-L} on \textbf{Real Faces} and \textbf{Real World} splits, and also \textbf{Forget Quality} and \textbf{Model Utility} metrics. 

\section{Multimodal unlearning on QwenVL series}
\label{sec:qwen_results}
In addition to our experiments on the LLAVA series, we provide results for the QwenVL2-2B model in the table \ref{tab:qwen_results}. We use same hyperparameters as stated in \ref{sec:multimodal_unlearning}, except that we do not use the LoRA adapters.

\section{U-MIA and U-LIRA}
In this section, we provide details on evaluating unlearning methods using Unlearning Membership Inference Attack (U-MIA) algorithms. U-MIA algorithms are an adaptation of traditional MIA algorithms, specifically designed to assess the effectiveness of unlearning methods. The primary distinction between standard MIA and its unlearning counterpart lies in their objectives. Traditional MIA algorithms aim to determine whether a particular example was included in the training dataset of a model. In contrast, U-MIA algorithms are designed to detect whether a model was initially trained on a specific example and then subjected to an unlearning algorithm or if the model has never encountered the example at all.

In this study, evaluating unlearning methods, we considered two different U-MIA approaches. The first one is based on the original MIA introduced in \cite{shokri2017membership}. It assumes training a specific classifier which for any input example (x, y) will output the probability that object x was forgotten by the model. The second one exploits the LIRA approach introduced in \cite{carlini2022membershipinferenceattacksprinciples}. It is based on the Likelihood-ratio Test between hypotheses H1 and H2, where H1: object x comes from Q1 (forget distribution) and H2: x comes from Q2 (holdout distribution). 

\label{sec:umia_ulira}

